%% file: main_tpami.tex
\documentclass[10pt,journal,compsoc]{IEEEtran}
\usepackage{times}
\usepackage{epsfig}
\usepackage{graphicx}
\usepackage{amsmath}
\usepackage{amssymb}
\usepackage{tabularx}
\usepackage{multirow}
\usepackage{color}

\usepackage[nocompress]{cite}
\usepackage{amsmath}

\begin{document}

\title{Referring Segmentation in Images and Videos with Cross-Modal Self-Attention Network}

\author{Linwei Ye,~
	Mrigank Rochan,~
	Zhi Liu,~\IEEEmembership{Senior Member,~IEEE,}~
        Xiaoqin Zhang
        and~Yang Wang
\IEEEcompsocitemizethanks{\IEEEcompsocthanksitem
L. Ye is now with College of Computer Science and Artificial Intelligence, Wenzhou University, Wenzhou, 325035, China. He was with the Department of Computer Science, University of Manitoba, Winnipeg, MB R3T 2N2, Canada (e-mail: yel3@cs.umanitoba.ca). 
 
\IEEEcompsocthanksitem
M. Rochan and Y. Wang are with the Department of Computer Science, University of Manitoba, Winnipeg, MB R3T 2N2, Canada (e-mail: \{mrochan, ywang\}@cs.umanitoba.ca).

\IEEEcompsocthanksitem Z. Liu is with Shanghai Institute for Advanced Communication and Data Science, and School of Communication and Information Engineering, Shanghai University, Shanghai 200444, China (e-mail: liuzhi@staff.shu.edu.cn).
\IEEEcompsocthanksitem X. Zhang is with College of Computer Science and Artificial Intelligence, Wenzhou University, Wenzhou, 325035, China. (email: zhangxiaoqinnan@gmail.com)
}
\thanks{Corresponding authors: Zhi Liu and Yang Wang.}}



\IEEEtitleabstractindextext{
\begin{abstract}
\input abstract.tex
\end{abstract}

}

\maketitle

\IEEEdisplaynontitleabstractindextext

%
\IEEEpeerreviewmaketitle

\input intro.tex

\input related.tex

\input approach.tex

\input experiment.tex

\input conclude.tex

\section*{Acknowledgment} 
This work was supported in part by the NSERC, in part by the National Natural Science Foundation of China [Grant Nos. 61771301, 61922064, U2033210], in part by the Zhejiang Provincial Natural Science Foundation [Grant No. LR17F030001] and in part by the FoS research chair program and the GETS program at the University of Manitoba.

\bibliographystyle{IEEEtran}
\bibliography{J_ref}

\end{document}

%% file: abstract.tex
We consider the problem of referring segmentation in images and videos with natural language. Given an input image (or video) and a referring expression, the goal is to segment the entity referred by the expression in the image or video.
In this paper, we propose a cross-modal self-attention (CMSA) module to utilize fine details of individual words and the input image or video, which effectively captures the long-range dependencies between linguistic and visual features. Our model can adaptively focus on informative words in the referring expression and important regions in the visual input. We further propose a gated multi-level fusion (GMLF) module to selectively integrate self-attentive cross-modal features corresponding to different levels of visual features. This module controls the feature fusion of information flow of features at different levels with high-level and low-level semantic information related to different attentive words. Besides, we introduce cross-frame self-attention (CFSA) module to effectively integrate temporal information in consecutive frames which extends our method in the case of referring segmentation in videos. Experiments on benchmark datasets of four referring image datasets and two actor and action video segmentation datasets consistently demonstrate that our proposed approach outperforms existing state-of-the-art methods.

%% file: intro.tex
\IEEEraisesectionheading{\section{Introduction}\label{sec:intro}}

\begin{figure}[htb]
  \centering
  \includegraphics[width=7.3cm]{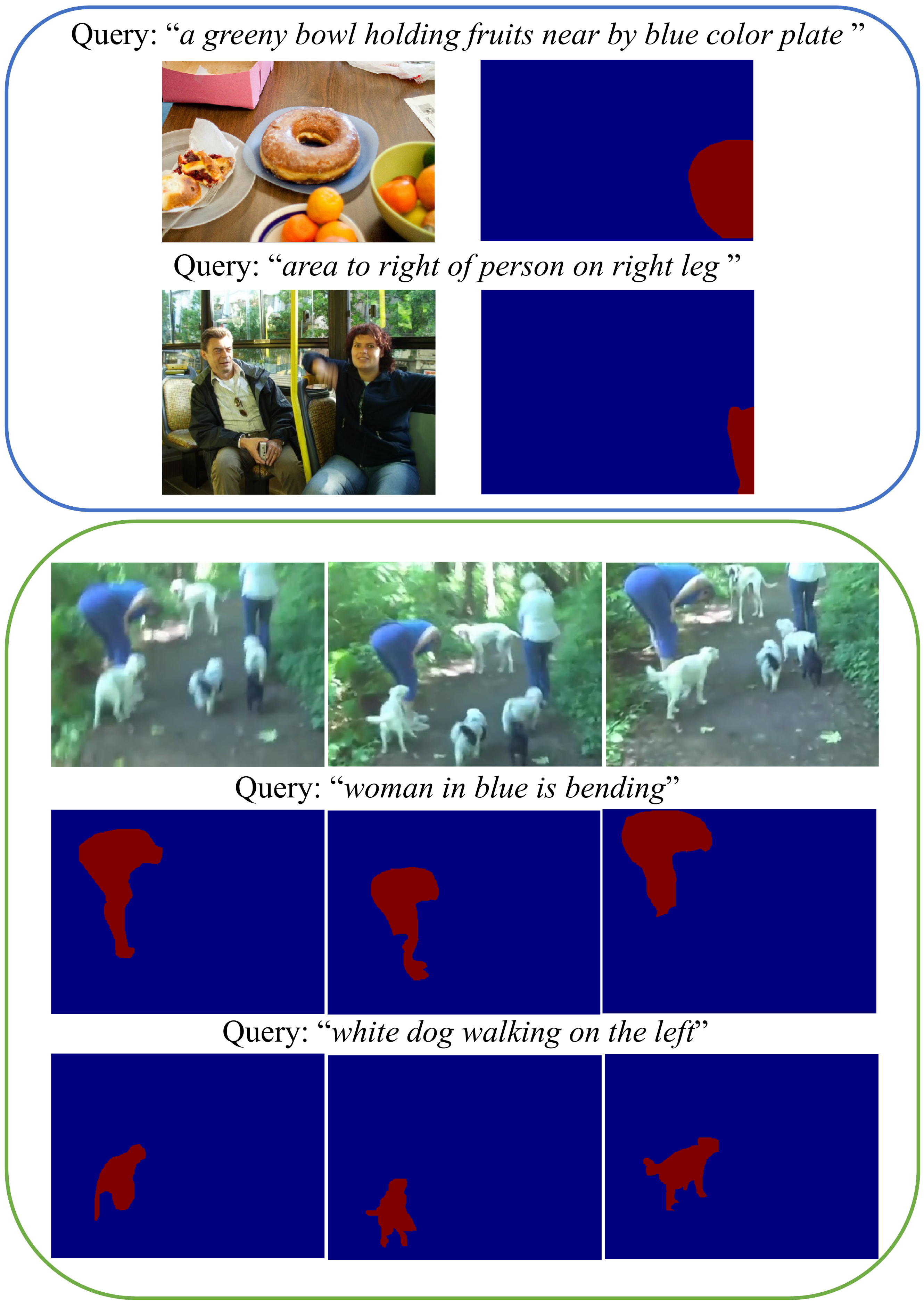}
\caption{Referring segmentation in images and videos. Given an image (top two examples) or a clip of a video (bottom example) with a referring expression, the segmentation mask is produced according to the corresponding expression query.} 
\label{fig:intro_im} 
\end{figure}

\IEEEPARstart{L}{anguage} is a natural media of communication between human and computers. In the computer vision community, there has been a lot of work at the intersection of vision and language. Many novel and interesting tasks of scene understanding are explored to combine linguistic and visual data simultaneously, such as image captioning~\cite{xu2015show,you2016image,ye2018attentive}, visual question answering~\cite{antol2015vqa,yang2016stacked} and visual grounding~\cite{hu2016natural,yu2018mattnet,deng2018visual}. In this paper, we investigate the referring segmentation problem in images and videos. Given an input image (or video) and a referring expression in natural language, the goal is to produce a segmentation mask of the referred entity in the image (or frames of the video). A reliable solution of this problem will provide a natural way for human-machine interactions. Referring segmentation is a challenging problem at the intersection of computer vision and natural language processing. Learning correlations between visual and linguistic modalities in the spatial regions is necessary to accurately identify the referred entity and produce the corresponding segmentation mask.

Referring segmentation is different from other well-explored segmentation tasks that assume a pre-defined category set, including semantic segmentation~\cite{noh2015learning,chen2016deeplab}, instance segmentation~\cite{he2017mask}, or biology-inspired models based on the human visual cognition system for saliency detection~\cite{cheng2014global,ye2017salient}. The object-of-interest in referring segmentation has to be implicitly inferred according to the query expression in the natural language form.

\begin{figure}[htb]
  \centering
  \includegraphics[width=8.4cm]{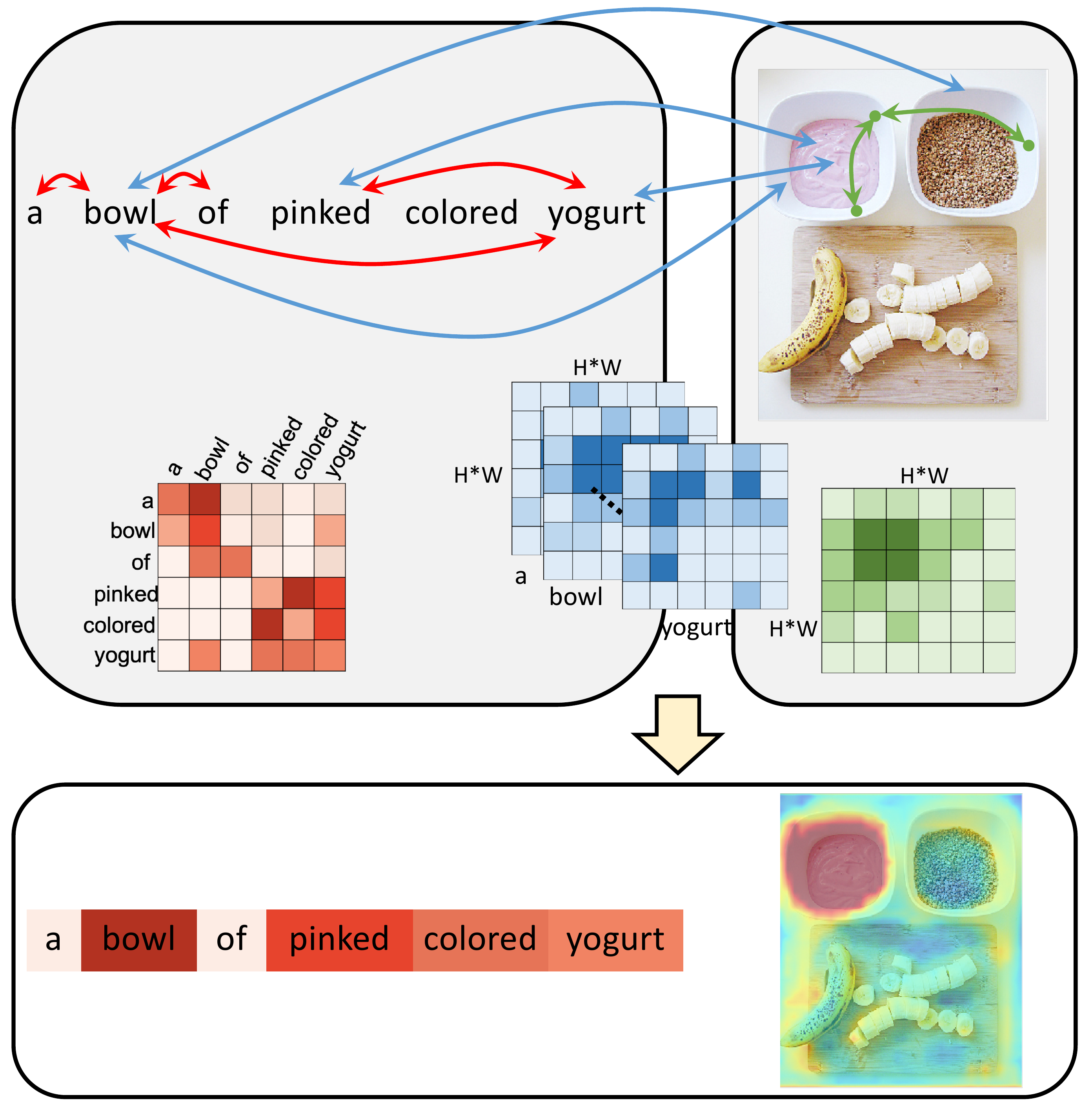}
\caption{(Best viewed in color) Illustration of our cross-modal self-attention mechanism. It is composed of three joint operations: self-attention over language (shown in red), self-attention over image representation (shown in green), and cross-modal attention between language and image (shown in blue). The visualizations of linguistic and spatial feature representations (in bottom row) show that the proposed model can focus on specific key words in the language and spatial regions in the image that are necessary to produce precise referring segmentation masks.} 
	
\label{fig:intro} 
\end{figure}

Fig.~\ref{fig:intro_im} illustrates some examples for this task. Given an image or a clip of a video with an expression query, the goal is to produce a pixel-level segmentation mask specified by the query shown on the top of the corresponding image/video frames. We can observe that the referring expression is not limited to specifying object categories (e.g. ``person'', ``dog''). It can take any free form of language descriptions which may contain appearance attributes (e.g. ``greeny''), actions (e.g. ``holding'', ``walking'') and relative relationships (e.g. ``near'', ``right of''), etc. It is also noteworthy that the referred entity can be more flexibly represented than traditional segmentation tasks. For instance, in the second example of Fig.~\ref{fig:intro_im}, the referred regions can not be represented by a simple word. Instead, the referred entity can be a certain part of the image with relative relationships to other explicit objects (e.g., ``right leg'').

Several existing works have been proposed to tackle this problem. A popular approach (e.g. \cite{hu2016segmentation,li2018referring,shi2018key,gavrilyuk2018actor}) in this area is to use convolutional neural network (CNN) to extract visual feature, and separably adopt recurrent neural network (RNN) or another CNN to  represent  the referring expression. The resultant visual and linguistic representations are then combined by concatenation~\cite{hu2016segmentation,li2018referring}, visual context~\cite{shi2018key} or dynamic filter~\cite{gavrilyuk2018actor} to produce the final pixel-wise segmentation result. However, the limitation of this approach is that the language encoding module directly encodes the referring expression as a whole and may ignore fine details of some individual words in the referring expression, which are important to identify the referred entities and produce an accurate segmentation mask.

Some previous works (e.g.~\cite{liu2017recurrent,margffoy2018dynamic}) focus on learning multimodal interaction in a sequential manner to process each word in the referring expression. Specifically, the visual feature is sequentially merged with every encoding word from the output of LSTM-based~\cite{hochreiter1997long} language model at each step to infer a multimodal feature representation. But the multimodal interaction in these works only considers the linguistic and visual information individually within their local contexts. It may not sufficiently capture global interaction information essential for semantic understanding and segmentation.

To address the limitations of aforementioned methods, we propose a \textit{cross-modal self-attention (CMSA)} module to effectively learn long-range dependencies from multimodal features that represent both visual and linguistic information. Our model can adaptively focus on important regions in the image and informative keywords in the language description. Fig.~\ref{fig:intro} shows an example that illustrates the cross-modal self-attention module, where the correlations among words in the language and regions in the image are presented. We further develop a \textit{gated multi-level fusion (GMLF)} module to refine the segmentation mask of the referred entity. The gated fusion module is designed to selectively leverage multi-level self-attentive features which contain diverse visual semantic information related to different attentive words. In addition, we extend self-attention mechanism into the temporal domain to model dependencies across frames in a video clip. The proposed \textit{cross-frame self-attention (CFSA)} module can exploit temporary correlation information from a clip of the video to generate a robust representation for each frame.

In summary, our main contributions include: (1) A cross-modal self-attention (CMSA) module for referring segmentation. Our module effectively captures the long-range dependencies between linguistic and visual contexts to produce a robust multimodal feature representation for the task. (2) A gated multi-level fusion (GMLF) module to selectively integrate multi-level self-attentive features and effectively capture fine details for producing precise segmentation masks. (3) A cross-frame self-attention (CFSA) module for extracting effective feature information from a clip of the video for the current frame. (4) An extensive empirical study on four referring image segmentation benchmark datasets and two actor and action video segmentation datasets. The experimental results demonstrate that our proposed method achieves superior performance compared with state-of-the-art methods.

A preliminary version of this work has appeared as \cite{ye2019cross}. Compared with \cite{ye2019cross}, we improve the experimental results of our model by introducing a word attention-aware pooling module that extracts a more robust multimodal feature of individual word for an expression sentence. (2) The work in \cite{ye2019cross} only considers referring segmentation in images. In this paper, we also consider referring segmentation in videos. In particular, we propose a cross-frame self-attention (CFSA) module that captures the correlation between frames in a video clip. This module can be used to build more robust visual feature of each frame and improve the performance of referring segmentation in videos. (3) We perform more extensive experimental evaluations and more comprehensive review of related works. 

The rest of this paper is organized as follows. We first introduce the related work in Sec.~\ref{sec:related}. Then we describe the proposed approach with detailed descriptions for multimodal feature generation, cross-modal self-attention, gated multi-level fusion and cross-frame self-attention in Sec.~\ref{sec:approach}. Sec.~\ref{sec:experiment} presents experimental setup such as implementation details and datasets, and extensive experimental results for performance evaluation and comparison. Finally, conclusions are drawn in Sec.~\ref{sec:conclude}.

%% file: related.tex
\section{Related Work}
\label{sec:related}
%

In this section, we review several lines of research related to our work  in the fields of  attention, semantic segmentation, referring image localization and segmentation, actor and action video segmentation and referring video localization and segmentation.

\subsection{Attention}
Attention mechanism has been shown to be a powerful technique in deep learning models and has been widely used in various tasks in natural language processing~\cite{bahdanau2014neural,vaswani2017attention} to capture keywords for context. In the multimodal tasks, word attention has been used to re-weight the importance of image regions for image caption generation~\cite{xu2015show}, image question answering~\cite{yang2016stacked} and referring image segmentation~\cite{shi2018key}. In addition, attention is also used for modeling subject, relationship and object~\cite{hu2017modeling} and for referring relationship comprehension~\cite{yu2018mattnet}. The diverse attentions of query, image and objects are calculated separately and then accumulated circularly for visual grounding in~\cite{deng2018visual}.

Self-attention~\cite{vaswani2017attention} is proposed to attend a word to all other words for learning relations in the input sequence. It significantly improves the performance for machine translation. This technique is also introduced in videos to capture long-term dependencies across temporal frames~\cite{wang2018non}. 
Inspired by bidirectional encoder representations from transformers (BERT)~\cite{devlin2018bert}, a VisualBERT model is proposed by~\cite{li2019visualbert} to simultaneously integrate BERT and self-attention mechanism with bounding boxes for several vision-and-language tasks.  Another BERT-like model ViLBERT~\cite{lu2019vilbert} addresses vision and language inputs in two separate streams and then use co-attentional transformer layers to learn a joint task-agnostic representation. 
Different from these works, we propose a cross-modal self-attention module to bridge attentions across language and vision. Our network can achieve fine-grained interaction between each region in the visual features and each word in the linguistic features.

\subsection{Semantic segmentation}
Semantic segmentation has achieved great advances in recent years. Fully convolutional networks (FCN)~\cite{long2015fully} take advantage of fully convolutional layers to train a segmentation model in an end-to-end way by replacing fully connected layers in CNN with convolutional layers. In order to alleviate the down-sampling issue,  deconvolution network~\cite{noh2015learning} that consists of  deconvolution and unpooling layers are appended to the CNN for detailed structures and scales of the objects. 
DeepLab~\cite{chen2016deeplab} further adopts dilated convolution to enlarge the receptive field and uses atrous spatial pyramid pooling for multi-scale segmentation. An improved pyramid pooling module~\cite{zhao2017pyramid} further enhances the use of multi-scale structure. These methods enlarges the semantic context to produce precise segmentation masks. Lower level features are explored to bring more detailed information to complement high-level features for generating more accurate segmentation masks~\cite{badrinarayanan2015segnet,long2015fully,noh2015learning}.

\subsection{Referring image localization and segmentation}
Referring image localization aims to localize specific objects in an image according to the description of a referring expression. It has been explored in natural language object retrieval~\cite{hu2016natural} and modeling relationship~\cite{hu2017modeling,yu2018mattnet}. In order to obtain a more precise result,  referring image segmentation is proposed to produce a segmentation mask instead of a bounding box. This problem was first introduced in \cite{hu2016segmentation}, where CNN and LSTM are used to extract visual and linguistic features, respectively. They are then concatenated together for spatial mask prediction. To better achieve word-to-image interaction, \cite{liu2017recurrent} directly combines visual features with each word feature from a language LSTM to recurrently refine segmentation results. Dynamic filter~\cite{margffoy2018dynamic} for each word further enhances this interaction. In~\cite{shi2018key}, word attention is incorporated in the image regions to model key-word-aware context. Low-level visual features are also exploited for this task in~\cite{li2018referring}, where Convolutional LSTM (ConvLSTM)~\cite{xingjian2015convolutional} progressively refines segmentation masks from high-level to low-level features sequentially. In this paper, we propose to adaptively integrate multi-level self-attentive features by the gated fusion module.

\subsection{Actor and action video segmentation}
The task of actor and action video segmentation pairs actor and its action to infer reasonable activities with the consideration of the main object. In~\cite{xu2015can}, a multi-layer conditional random field model is introduced to use supervoxel segmentation and features from a video for multiple-label cases in the Actor-Action Dataset. A multi-task detection method is proposed in~\cite{kalogeiton2017joint} to explore object-action relationships of  objects and their performing actions in a video from joint learning. Their work can be further extended to produce pixel-wise segmentation masks from bounding boxes. However, these video segmentation methods have a limited actor-action class space due to the demand of a pre-defined set of actor-action pairs. The recent advance of word embedding shows potential to enlarge a fixed class vocabulary to rich and flexible language descriptions. Several works have supplemented existing video datasets with linguistic sentences. A person search model utilizes a spatio-temporal human detection model for candidate tubes and multimodal feature embedding methods to retrieval person with person descriptions~\cite{yamaguchi2017spatio}. 

\subsection{Referring video localization and segmentation}
Conventionally, a target object needs to be specified in the first frame of a video by a pixel-level mask or scribbles for guiding video object segmentation in the remaining frames. Referring video localization and segmentation is a relatively new task. It uses referring expressions to identify the object of interest in videos. A language grounding model with temporal consistency and a static segmentation model are introduced by~\cite{khoreva2018video} to solve referring video segmentation problem. In~\cite{li2017tracking}, object localization and tracking is specified by a natural language query rather than specifying the target in the first frame of a video by a bounding box. A closely related work to this paper is~\cite{gavrilyuk2018actor}, which provides sentences for actor-action pairs and proposes an encoder-decoder architecture with dynamic filters. Recently, an asymmetric cross-guided attention network~\cite{wang2019asymmetric} leverages cross-modality attention mechanism to implement feature interaction.


%% file: approach.tex
\begin{figure*}[ht]
\begin{center}
\includegraphics[width=0.98\textwidth]{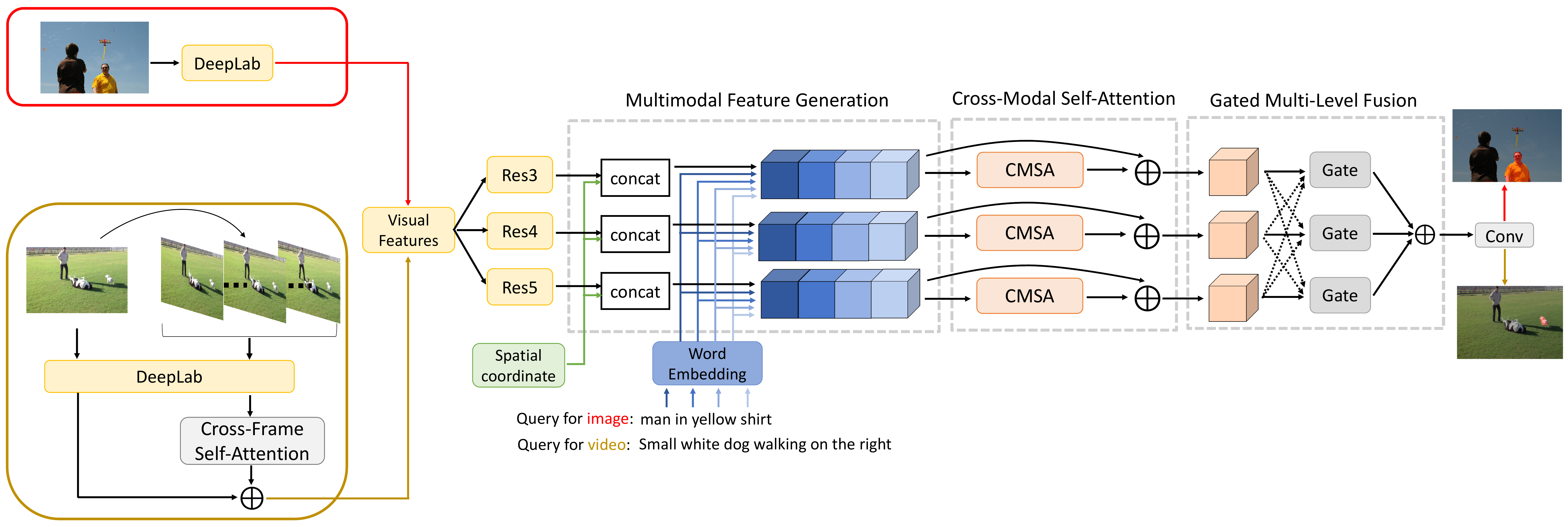}
\end{center}
   \caption{An overview of our approach. The proposed model consists of three components including multimodal feature generation, cross-modal self-attention (CMSA) and a gated multi-level fusion. Multimodal feature generation are constructed from the visual feature, the spatial coordinate feature and the language feature for each word. For referring segmentation in video, a clip of adjacent frames is used to extract temporal correlation feature by cross-frame self-attention (CFSA) for producing visual feature of the current frame as shown in the gold box. Then the multimodal feature at each level is fed to a cross-modal self-attention module to build long-range dependencies across individual words and spatial regions. Finally, the gated multi-level fusion module combines the features from different levels to produce the final segmentation mask. The red arrows and gold arrows show the input and output for the image and video, respectively.}
\label{fig:overview}
\end{figure*}

\section{The Proposed Model}\label{sec:approach}

The overall architecture of our model is shown in Fig.~\ref{fig:overview}. The input consists of an image or a clip of a video, with a referring expression as the query. We first use a CNN to extract visual feature maps at different levels from the input image (red box at top left in Fig.~\ref{fig:overview}) or a clip of a video enhanced by the cross-frame self-attention module (gold box at bottom left in Fig.~\ref{fig:overview}). Each word in the referring expression is represented as a vector of word embedding. Every word vector is then appended to the visual feature map to produce a multimodal feature map. Thus, there is a multimodal feature map for each word in the referring expression. We then introduce self-attention~\cite{vaswani2017attention} mechanism to combine the feature maps of different words into a cross-modal self-attentive feature map. The self-attentive feature map captures rich information and long-range dependencies of both linguistic and visual information of the inputs. In the end, the self-attentive features from multiple levels are combined via a gating mechanism to produce the final features used for generating the segmentation output. For referring segmentation in videos, the input consists of a video and a referring expression (bottom left in Fig.~\ref{fig:overview}). In this case, we first use a cross-frame self-attention module to capture the temporal correlations between frames within a video clip. We then use the output of this module to build better visual features for frames in a video. Then we apply the remaining pipeline in the architecture to produce the segmentation mask in each frame of the video.

Our model is motivated by several observations. First of all, in order to solve referring segmentation, we typically require detailed information of certain individual words (e.g. words like ``left'', ``right''). Previous works~(e.g.~\cite{hu2016segmentation,li2018referring,shi2018key}) take word vectors as inputs and use LSTM to produce a vector representation of the entire referring expression. The vector representation of the entire  referring expression is then combined with the visual features for referring image segmentation. The potential limitation of this technique is that the vector representation produced by LSTM captures the meaning of the entire referring expression while missing sufficiently detailed information of some individual words needed for the referring image segmentation task. Our model addresses this issue by maintaining word vector for each word in the entire referring expression instead of using LSTM. Therefore, it can better capture more detailed word-level information. Secondly, some previous works~(e.g.~\cite{liu2017recurrent,margffoy2018dynamic}) process each word in the referring expression and concatenate it with visual features to infer the referred object in a sequential order using a recurrent network. The limitation of these methods is that these methods only look at each spatial region of feature maps locally and lack the interaction over long-range spatial regions in global context which is essential for semantic understanding and segmentation. In contrast, our model uses a cross-modal self-attention module that can effectively model long-range dependencies between linguistic and visual modalities. Thirdly, different from~\cite{li2018referring} which adopts ConvLSTM to refine segmentation with multi-scale visual features sequentially, the proposed method employs a novel gated fusion module for combining multi-level self-attentive features. 
Lastly, we introduce the cross-frame self-attention module to build long-range dependencies across frames in the video. It learns temporal correlations from adjacent frames for the current frame and extends our model to video domain.

\subsection{Multimodal Feature Generation}\label{sec:mfg}
We first introduce the generic modules of the proposed model by taking an image as input for simplification, then we will present details of the cross-frame self-attention module and illustrate how to extend our model to video domain. Let $I$ denote an input image and $N$ represent the length of a referring expression. Then each individual word in the referring expression can be denoted as $w_n, n\in{1, 2,...,N}$. We use a backbone CNN network to extract visual features from the input image $I$. The feature map is extracted from a specific CNN layer and represented as $V\in\mathbb{R}^{H\times W\times C_v}$, where $H$, $W$ and $C_v$ are the dimensions of height, width and feature channel, respectively. For ease of presentation, we only use features extracted from one particular CNN layer for now. Later in Sec.~\ref{sec:gf}, we present an extension of our method that uses multi-level features from multiple CNN layers for gated fusion. 

For the language description with $N$ words, each word $w_n$ is encoded as a one-hot vector to project it into a compact word embedding represented as $\mathbf{e}_n\in\mathbb{R}^{C_l}$ by a lookup table. Different from previous methods~\cite{hu2016segmentation,li2018referring,shi2018key} that simply apply LSTM to process the word vectors sequentially and take the output hidden layer to encode the entire description sentence with a sentence vector, we propose to keep the individual word vector and introduce a cross-modal self-attention module to capture long-range correlations between these words and spatial regions in the image. More details will be presented in Sec.~\ref{sec:msa}.

In addition to visual features and word vectors, spatial coordinate features have also been shown to be useful for referring image segmentation~\cite{hu2016segmentation,li2018referring,liu2017recurrent}. Following prior works, we define an $8$-D spatial coordinate feature map at each spatial position using the implementation in~\cite{liu2017recurrent}. Specifically, the first $3$-dimensions of the spatial coordinate feature map encode the normalized horizontal positions. The next $3$-dimensions encode normalized vertical positions. The last $2$-dimensions encode the normalized width and height information of the image.

After obtaining the visual, linguistic and spatial features, we finally construct a joint multimodal feature representation at each spatial position for each word by concatenating the visual features, word vectors, and spatial coordinate features. Let $p$ be a spatial position in the feature map $V$, i.e. $p\in\{1,2,..., H\times W\}$. We use $\mathbf{v}_p\in\mathbb{R}^{C_v}$ to denote the ``slice'' of the visual feature vector at a specific spatial location $p$. The spatial coordinate feature of the position $p$ is denoted as $\mathbf{s}_p\in\mathbb{R}^8$. Thus the multimodal feature $\mathbf{f}_{pn}$ can be defined corresponding to the position $p$ and the $n$-th word as follows:
\begin{equation}
\mathbf{f}_{pn} = Concat\left(\frac{\mathbf{v}_p}{||\mathbf{v}_p||_2}, \frac{\mathbf{e}_n}{||\mathbf{e}_n||_2}, \mathbf{s}_p\right)
\end{equation}
where $||\cdot||_2$ denotes the $L_2$ norm of a vector and $Concat(\cdot)$ denotes the concatenation of several input vectors. The multimodal feature vector $\mathbf{f}_{pn}$ encodes information about the combination of a specific position $p$ in the image and the $n$-th word $w_{n}$ in the referring expression with a total dimension of $(C_v+C_l+8)$. We use $F=\{\mathbf{f}_{pn}: \forall p, \forall n\}$ to represent the collection of features $\mathbf{f}_{pn}$ for different spatial positions and words. The final dimension of $F$ is $N\times H\times W\times (C_v+C_l+8)$.

\subsection{Cross-Modal Self-Attention}\label{sec:msa}
The generated multimodal feature $F$ maintains rich and fine features for each spatial position and word. However, it is quite huge to further process directly and may contain a lot of redundant information. Additionally, the size of $F$ is variable depending on the number of words in the referring expression. It is difficult to directly exploit $F$ to produce the segmentation output. In recent years, the attention mechanism~\cite{hu2017modeling,shi2018key,vaswani2017attention,xu2015show,yu2018mattnet} has been shown to be a powerful technique that can capture important information from raw features in either linguistic or visual representation. Different from above works, we propose a cross-modal self-attention module to jointly exploit attentions over detailed ultimodal features. In particular, inspired by the success of self-attention~\cite{vaswani2017attention,wang2018non}, our designed cross-modal self-attention module can capture long-range dependencies in three aspects including self-attention over language, self-attention over image representation and cross-modal attention between the words in a referring expression and different spatial positions in the input image. The proposed module takes $F$ as the input and produces a feature map that summarizes $F$ after learning the correlation between the language expression and the visual context.

Given a multimodal feature vector $\mathbf{f}_{pn}$, the cross-modal self-attention module first produces a set of query, key and value pair by linear transformations as $\mathbf{q}_{pn}=W_q\mathbf{f}_{pn}$ , $\mathbf{k}_{pn}=W_k\mathbf{f}_{pn}$ and  $\mathbf{v}_{pn}=W_v\mathbf{f}_{pn}$ at each spatial position $p$ and the $n$-th word, where $\{W_q, W_k, W_v\}$ are part of the model parameters to be learned. Each query, key and value is reduced from the high dimension of multimodal features to the dimension of $512$ in our implementation, i.e. $W_q,W_k,W_v\in\mathbb{R}^{512\times (C_v+C_l+8)}$, for computation efficiency.

We then compute the cross-modal self-attentive feature $\widehat{\mathbf{v}}_{pn}$ as follows: 
\begin{eqnarray}
  && \hspace{-20pt} \widehat{\mathbf{v}}_{pn} = \sum_{p'}\sum_{n'}a_{p,n,p',n'} \mathbf{v}_{p'n'}, \ \ \ \textrm{where }\label{eq:self1}\\
  && \hspace{-20pt} a_{p,n,p',n'}=\textrm{Softmax}(\mathbf{q}_{p'n'}^T \mathbf{k}_{pn})\label{eq:self2}
\end{eqnarray}
where $a_{p,n,p',n'}$ is the attention score that takes into account of the correlation between $(p,n)$ and any other combinations of spatial position and word $(p',n')$.

Then $\widehat{\mathbf{v}}_{pn}$ is transformed back to the same dimension as $\mathbf{f}_{pn}$ via another linear layer and is added element-wise with $\mathbf{f}_{pn}$ to form a residual connection as $\widehat{\mathbf{f}}_{pn} = W_{\widehat{v}}\widehat{\mathbf{v}}_{pn}+ \mathbf{f}_{pn}$. This allows the insertion of this module into to the backbone network without breaking its behavior~\cite{he2016deep}.

\noindent{\bf Word attention-aware pooling:}
In order to collect the attention distribution of each word and produce a  single output feature map for segmentation, we propose the word attention-aware pooling to summarize attention scores by pooling separate multimodal feature of individual word to a feature representation for the whole expression sentence.

\begin{figure}[htb]
  \centering
  \includegraphics[width=0.5\textwidth]{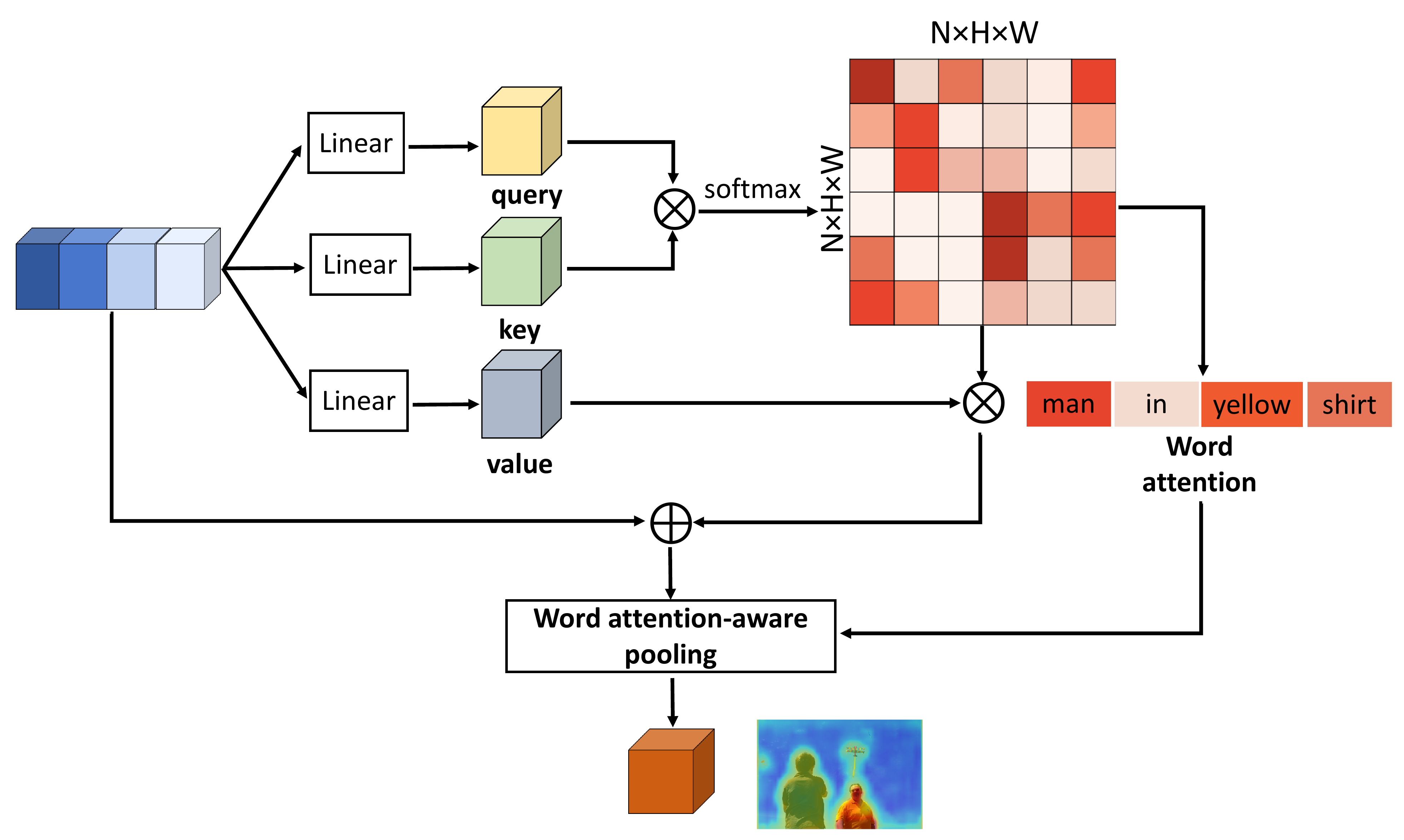}
    \caption{An illustration of the process of generating the cross-modal self-attentive (CMSA) feature from an image and a language expression (``man in yellow shirt''). We use $\otimes$ and $\oplus$ to denote matrix multiplication and element-wise summation, respectively. The softmax operation is performed over each row which indicates the attentions across each visual and language cell in the multimodal feature. The word attention-aware pooling summarizes word-level features for multimodal representation of the referring expression. We also visualize the internal linguistic and spatial representations. Please refer to Sec.~\ref{sec:result} and Sec.~\ref{sec:vis} for more details.}
\label{fig:nl}
\end{figure}

From the collection of the attention scores $\{a_{p,n,p',n'}: \forall p, \forall n, \forall p', \forall n'\}$ in Eq.~\ref{eq:self2}, we sum up all combinations for different pairs of $(p,n)$ and $(p',n')$ for each word $n$. As a result, we can get a vector of length $N$ as the accumulation of attention scores for all words.  The word attention score over all words in the expression  can be calculated by softmax operation as:
\begin{equation} 
a_{n} = \textrm{Softmax}(\sum_{p,p',n'}a_{p,n,p',n'})
\label{eq:w_att}
\end{equation}
Using Eq.~\ref{eq:w_att}, we can obtain word attentions from the intermediate result of the self-attention matrix and use this information to adaptively pool individual word feature vector together for the expression sentence. To be more specific, we multiply each multimodal feature vector $\widehat{\mathbf{f}}_{pn}$ by its corresponding word attention $a_{n}$ and accumulate $\widehat{\mathbf{f}}_{pn}$ over all words to result in the final feature representation:
\begin{equation}
    \widehat{\mathbf{f}}_{p} =  \sum_n a_{n} \widehat{\mathbf{f}}_{pn} \label{eq:final}
\end{equation}
The final feature map $\widehat{F}=\{\widehat{\mathbf{f}}_p: \forall p\}$ denotes the collection of $\widehat{\mathbf{f}}_{p}$ at all spatial positions, i.e. $\widehat{F}\in\mathbb{R}^{H\times W\times (C_v+C_l+8)}$.
Unlike previous methods that keep the entire input feature size~\cite{wang2018non} or use average-pooling to generate spatial feature map~\cite{ye2019cross}, the proposed word attention-aware pooling effectively reduces the size of input multimodal feature and fits different numbers of input words present in the sentence. It also obtains informative features by taking advantage of the intermediate result of the self-attention operation. The summarization of the attention score for each word help generate effective multimodal representation of the referring expression from word-level features. Fig.~\ref{fig:nl} illustrates the process of generating cross-modal self-attentive features.

\subsection{Gated Multi-Level Fusion}\label{sec:gf}
The feature representation $\widehat{F}$ obtained from Eq.~\ref{eq:final} is specific to a particular layer in CNN. Previous work~\cite{li2018referring} has shown that fusing features at multiple scales can improve the performance of referring image segmentation. In this section, we introduce a novel gated fusion technique to integrate multi-level features.

Let $\widehat{F^{(i)}}$ be the cross-modal self-attentive feature map at the $i$-th level. Following~\cite{li2018referring}, we use ResNet based DeepLab-101 as the backbone CNN and consider feature maps at three levels ($i=1,2,3$) corresponding to ResNet blocks $Res3$, $Res4$ and $Res5$. Let $C_{v_{i}}$ be the channel dimension of the visual feature map at the $i$-th level of the network. We use $\widehat{F^{(i)}}=\{\widehat{\mathbf{f}}^{(i)}_{p}: \forall p\}$ to indicate the collection of cross-modal self-attentive features $\widehat{\mathbf{f}}^{(i)}_{p}\in\mathbb{R}^{C_{v_{i}}+C_l+8}$ for different spatial locations corresponding to the $i$-th level. Our goal is to fuse the feature maps $\widehat{F^{(i)}}$ ($i=1,2,3$) to produce a fused feature map for producing the final segmentation output.
Note that the feature maps $\widehat{F^{(i)}}$  have different channel dimensions at different level $i$. At each level, we apply a $1\times 1$ convolutional layer to make the channel dimensions of different levels consistent and result in an output $X^{(i)}$.

For the $i$-th level, we use linear transformations for $X^{(i)}$ to generate a memory gate $m^{i}$ and a reset gate $r^{i}$ ($r^{i}, m^{i}\in\mathbb{R}^{H_i\times W_i}$), respectively. These gates play similar roles to the gates in LSTM. Different from stage-wise memory updates~\cite{cho2014learning,hochreiter1997long}, the computation of gates at each level is decoupled from other levels. The gates at each level control how much the feature at each level contributes to the final fused feature. Each level also has its own contextual controller $G^{i}$ which modulates the information flow from other levels to the $i$-th level. This process can be summarized as: \begin{align}
\label{eq:mgi}
\begin{split}
\hspace{-10pt}    G^{i} & = (1 - m^{i}) \odot X^{i} + \sum_{j\in \{1,2,3\} \setminus \{i\}}\gamma^{j} m^{j} \odot X^{j}\\
\hspace{-10pt}    F^{i}_o & = r^{i} \odot \textrm{tanh}(G^{i}) + (1-r^{i}) \odot X^i, \ \ \forall i \in \{1,2,3\}\\
\end{split}
\end{align}
where $\odot$ denotes Hadamard product. $\gamma^j$ is a learnable parameter to adjust the relative ratio of the memory gate which controls information flow of features from different levels $j$ combined to the current level $i$. 

In order to obtain the segmentation mask, we aggregate the feature maps $F^i_{o}$ from the three levels and apply a $3\times 3$ convolutional layer followed by the sigmoid function. This sequence of operations outputs a probability map ($P$) indicating the likelihood of each pixel being the foreground in the segmentation mask, i.e.:
\begin{equation}
P = \sigma\left(\mathbb{C}_{3\times 3} \left(\sum_{i=1}^{3} F_{o}^i\right)\right)
\end{equation}
where $\sigma(\cdot)$ and $\mathbb{C}_{3\times 3}$ denote the sigmoid and $3\times 3$ convolution operation, respectively. A binary cross-entropy loss function is defined on the predicted output and the ground-truth segmentation mask $Y$ as follows:
\begin{equation}
    L = - \frac{1}{\Omega}\sum_{m=1}^{\Omega}(Y(m)\log P(m)+(1-Y(m))\log(1-P(m)))\label{eq:final_loss}
\end{equation}
where $\Omega$ is the whole set of pixels in the image and $m$ is m-th pixel in it. We use the Adam algorithm~\cite{kingma2014adam} to optimize the loss in Eq.~\ref{eq:final_loss}.

\subsection{Cross-Frame Self-Attention}\label{sec:cfsa}

The task of actor and action video segmentation from a sentence extends a fixed number of actor and action pairs to an open set of vocabulary with rich language descriptions. It is similar to the referring image segmentation, but requires producing a segmentation mask in frames of a video according to the referring query. We use this task to demonstrate the referring segmentation in videos. We propose a cross-frame self-attention module to learn temporal correlations from adjacent  frames for  the current frame. For an input frame $T$, we extract an adjacent clip of the current frame with the interval of $\tau$.  The CNN backbone is then used to extract features for every frame in the clip and result in a set of features as $\{V_t\}_{T-\tau}^{T+\tau}$. Similar to the implementation in Eq.~\ref{eq:self1} and Eq.~\ref{eq:self2}, let $\mathbf{v}_{t,p}\in\mathbb{R}^{C_v}$ be a feature vector in a frame $t$ and at a specific spatial position $p$. The cross-frame self-attention feature $\widehat{\mathbf{v}}_{tp}$ can be obtained by:
\begin{eqnarray}
  && \hspace{-20pt} \widehat{\mathbf{v}}_{tp} = \sum_{p'}\sum_{t'}a_{t,p,t',p'} \mathbf{v}_{t'p'}, \ \ \ \textrm{where }\label{eq:self3}\\
  && \hspace{-20pt} a_{t,p,t',p'}=\textrm{Softmax}(\mathbf{q}_{t'p'}^T \mathbf{k}_{tp})\label{eq:self4}
\end{eqnarray}
where $a_{t,p,t',p'}$ is the attention score that represents the correlation between $(t,n)$ and any other combinations of frame and spatial position $(t',p')$.
Attention-aware pooling of frames is applied over frames to generate the temporal correlation feature $\widehat{\mathbf{v}}_{p}$ as follows:

\begin{equation}
    \widehat{\mathbf{v}}_{p} =  \sum_t \widehat{\mathbf{v}}_{tp} ( \textrm{Softmax}(\sum_{t',p,p'}a_{t,p,t',p'}))
\label{eq:final_cfsa}
\end{equation}
The generated feature map $\widehat{V}=\{\widehat{\mathbf{v}}_p: \forall p\}$ collects $\widehat{\mathbf{v}}_{p}$ at all spatial positions. Then the temporal correlation feature $\widehat{V}$ is used to update the visual feature $V_t$ at the frame $t$ as: $V_t := V_t + \widehat{V}$. The final visual feature $V_t$  incorporates correlations learned at different positions across all frames in the clip to better capture temporal information for the referring segmentation in videos. Finally, $V_t$ can be seen as the visual feature $V$ in Sec.~\ref{sec:mfg} to combine with word vectors and spatial features for multimodal feature, and is followed by cross-modal self-attention module in Sec.~\ref{sec:msa} and gated multi-level fusion in Sec.~\ref{sec:gf} for producing segmentation masks.

%% file: experiment.tex
\section{Experiments}\label{sec:experiment}
In this section, we first introduce the experimental setup including implementation details, experimental datasets and evaluation metrics in Sec.~\ref{sec:expsetup}. Then we perform detailed ablation experiments to demonstrate the advantage and the relative contribution of each component of the proposed approach in Sec.~\ref{sec:ab}. Comprehensive quantitative and qualitative results of our method  are evaluated and compared with other state-of-the-art methods in Sec.~\ref{sec:result}. We also provide  visualization and failure cases to help gain insights of our model in Sec.~\ref{sec:vis}.

\subsection{Experimental Setup}\label{sec:expsetup}
\subsubsection{Implementation details}
Following previous contemporary work~\cite{li2018referring,liu2017recurrent, shi2018key}, we adopt DeepLab-101~\cite{chen2016deeplab} as our backbone network to extract visual features. The initial weights of the network are pretrained from Pascal VOC 2012 dataset~\cite{everingham2012pascal} for effective feature representation. The multi-level visual features are extracted from DeepLab-101 ResNet blocks $Res3,Res4,Res5$ as the inputs for multimodal features. 
We resize and zero-pad the spatial resolution of all input images as $320\times320$ and obtain the same width and height of feature maps at different levels due to the dilated convolution. The dimension used in $X^{(i)}$ for gated fusion is fixed to $500$.
In the language model, the maximum length of query expression are set to $20$ and we embed each word to a vector of $C_l=1000$ dimensions. 
For the segmentation in videos, we set $\tau=5$ to truncate totally $10$ adjacent frames besides a target frame for $CFSA$. 
The network is trained with Adam optimizer~\cite{kingma2014adam} with 
an initial learning rate of $2.5e^{-4}$ and weight decay of $5e^{-4}$. The learning rate is gradually decreased using the polynomial policy~\cite{chen2016deeplab} with the power of $0.9$. For fair comparisons, the final segmentation masks are refined by DenseCRF~\cite{krahenbuhl2011efficient} which is a common post-processing operation for precise segmentation results.

\subsubsection{Referring image segmentation datasets}
We perform extensive experiments on four referring image segmentation datasets to evaluation performance of our approach including UNC~\cite{yu2016modeling}, UNC+~\cite{yu2016modeling}, G-Ref~\cite{mao2016generation} and ReferIt~\cite{kazemzadeh2014referitgame}. The train, validation and test sets follow the setting at \cite{liu2017recurrent}.

The UNC dataset contains 19,994 images with 142,209 referring expressions for 50,000 objects. All images and expressions are collected from the MS COCO~\cite{lin2014microsoft} dataset interactively with a two-player game~\cite{kazemzadeh2014referitgame} where one player gives an expression description and anther player labels the corresponding referred region in the image. Two or more objects of the same object category can appear in each image. 

The UNC+ dataset is similar to the UNC dataset. but with a restriction that no location words are allowed in the referring expression. In this case, expressions regarding to referred objects totally depend on the appearance and scene context descriptions. 
It consists of 141,564 expressions for 49,856 objects in 19,992 images.

The G-Ref dataset is also collected based on MS COCO. It contains 104,560 expressions referring to 54,822 objects from 26,711 images. Annotations of this dataset come from Amazon Mechanical Turk instead of a two-player game. The average length of expressions is 8.4 words which is longer than that of other datasets (less than 4 words). It is more challenging to understand referring expressions for identifying the object of interest.

The ReferIt dataset is built upon the IAPR TC-12~\cite{escalante2010segmented} dataset. It has 130,525 expressions referring to 96,654 distinct object masks in 19,894 natural images. In addition to objects, it also contains mask annotations for stuff classes such as water, sky and ground.

\subsubsection{Actor and action video segmentation datasets}
Referring segmentation in videos is a relatively new problem. It also requires a referring expression to describe the expected object in the sequence of a video. Recently, Gavrilyuk \textit{et al.} \cite{gavrilyuk2018actor} complement segmentation masks in videos with referring expression descriptions in Actor-Action Dataset (A2D) dataset~\cite{xu2015can} and  Joint-annotated HMDB (JHMDB) dataset~\cite{jhuang2013towards} which are originally developed for video actor and action segmentation.

The A2D Sentences dataset is the largest general actor and action video segmentation dataset with pixel-level mask labels collected from Youtube database. It consists of 3,782 videos in total in which 3,036 and 746 videos are served
as train and test sets, respectively. There are seven classes of actors such as adult, baby and dog, and eight actions such as climbing, jumping, running and walking in the dataset. Referring expressions are augmented by \cite{gavrilyuk2018actor} with 6,656 sentences including 811 different nouns, 225 verbs and 189 adjectives. These linguistic annotations are labeled to discriminate each actor instance from multiple objects appeared in a video. Furthermore, it provides more diverse and finer granularities of descriptions where the actor class \textit{adult} in the referring sentences can be expressed as woman, man, person, player and so on. The average length of sentences contains 7.3 words.

The JHMDB Sentences dataset is a subset of HMDB51 database~\cite{kuehne2011hmdb}, containing 928 videos with 21 different actions involving brush hair, clap, sit, walk, etc. It also provides a 2D articulated human puppet for scale, pose, segmentation and a coarse viewpoint. The same strategy as for the A2D Sentences dataset is applied for referring expression annotations and ends up with 928 sentences including 158 different nouns, 53 verbs and 23 adjectives. It should be noted that all 928 videos in the JHMDB Sentences dataset are used for testing and evaluating the generalization ability of models.

\subsubsection{Evaluation metrics} 
In order to fairly compare with other methods, we follow previous work~\cite{li2018referring,liu2017recurrent, shi2018key} to use overall intersection-over-union $(Overall ~IoU)$ and Precision@X ($prec@X$) as the evaluation metrics. The $Overall ~IoU$ metric is a ratio between intersection and union of the predicted segmentation mask and the ground truth averaged over all test data. The $prec@X$ metric measures the percentage of test images with an IoU score higher than the threshold $X$, where $X \in\{ 0.5, 0.6,0.7,0.8,0.9\}$ in the experiments. For actor and action video segmentation, we additionally calculate $Mean ~IoU$ for balance between various sizes of target mask regions in different videos~\cite{gavrilyuk2018actor}. 

\input tab_ab_pool.tex
\input tab_ab_att.tex

\subsection{Ablation Study}\label{sec:ab}
We first perform ablation experiments on different variants of cross-modal self-attention module and then investigate the design choice of different attention methods and the relative contributions of multimodal feature representation and multi-level feature fusion of our proposed model.

\subsubsection{Variants of cross-modal self-attention module}
In order to evaluate the effectiveness of the proposed word attention-aware pooling method, we report the results of alternative methods that replace the word attention-aware pooling in our method with max pooling or average pooling. These methods use the same base model without multi-level feature fusion module or DenseCRF for postprocessing. As shown in Table~\ref{tab:ab_pool}, the word attention-aware pooling method consistently outperforms other two pooling methods. It demonstrates that the proposed word attention-aware pooling method can better collect the attention distribution of each word for the multimodal feature representation. The improvement is particularly significant on the G-Ref dataset. This also proves the power of the proposed word attention-aware pooling method for longer referring expressions since the G-Ref dataset has the longest average length of expressions among the four datasets.

\subsubsection{Attention methods}
We perform experiments to evaluate different attention methods in the multimodal feature representation. We alternatively use no attention, word attention, pixel attention and word-pixel pair attention by zeroing out the respective components in Eq.~\ref{eq:self1} to evaluate the influence of different attention combinations.

As shown in the top of Table~\ref{tab:ab_att}, attention methods achieve better segmentation performance than the no attention method. The proposed cross-modal self-attention outperforms all other attention methods significantly. This demonstrates that the attention method plays an important role in object segmentation and the language-to-vision correlation can be better learned together within our cross-modal self-attention method.

\subsubsection{Multimodal feature representation}
This experiment evaluates the effectiveness of the multimodal feature representation. Similar to the baselines, i.e. multimodal LSTM interaction in~\cite{liu2017recurrent} and convolution integration in~\cite{li2018referring}, we directly take the output of the $Res5$ of the network to test the performance of multimodal feature representation without the multi-level feature fusion. We use CMSA-W to denote the proposed method in Sec.~\ref{sec:msa}. In addition, a variant method CMSA-S which also uses the same cross-modal self-attentive feature, instead encodes the whole sentence to one single language vector by LSTM.

As shown in the middle four rows of Table~\ref{tab:ab_att}, the proposed cross-modal self-attentive feature based approaches achieve significantly better performance than other baselines, which manifests the advantage of our cross-modal self-attention module. Moreover, the word based method CMSA-W outperforms sentence based method CMSA-S. It reveals that encoding individual word in the referring expression can help capture detailed and useful information for multimodal feature representation.

\subsubsection{Multi-level feature fusion}
This experiment verifies the relative contribution of the proposed gated multi-level fusion module. Here we use our cross-modal self-attentive features as inputs and compare with several well-known network structures used for obtaining fine details of the segmentation mask, such as Deconvolution (Deconv)~\cite{noh2015learning} and Pyramid Pooling Module (PPM)~\cite{zhao2017pyramid} in semantic segmentation, and ConvLSTM~\cite{li2018referring} in referring image segmentation. 

In order to clearly understand the benefit of our multi-level feature fusion method, we also develop another self-gated method denoted as $CMSA+Gated$ that uses the same gate generation method in Sec.~\ref{sec:gf} to generate memory gates and directly multiply by its own features without interactions with features from other levels. As presented in the bottom five rows in Table~\ref{tab:ab_att}, the proposed gated multi-level fusion outperforms these other multi-scale feature methods which demonstrates the gated multi-level feature fusion module also benefits segmentation performance in terms of multi-level feature fusion.

\input tab_miou_unc.tex

\input tab_miou_unc+.tex

\subsection{Experimental Evaluation}\label{sec:result}
\subsubsection{Quantitative results of referring image segmentation}
We compare the referring image segmentation results of our method with other existing state-of-the-art approaches including recurrent multimodal interaction (RMI)~\cite{liu2017recurrent}, dynamic multimodal network  (DMN)~\cite{margffoy2018dynamic}, key-word-aware network (KWA)~\cite{shi2018key}, recurrent refinement networks (RRN)~\cite{li2018referring} and our preliminary work CMSA~\cite{ye2019cross}. We report both $prec@X$ and $Overall IoU$ at all four datasets in Table~\ref{tab:miou_unc},  Table~\ref{tab:miou_unc+},  Table~\ref{tab:miou_gref} and  Table~\ref{tab:miou_referit}, respectively.

Our proposed method consistently outperforms all other methods on all four datasets in terms of each threshold of $prec@X$ and $overall ~IoU$. The improvement is particularly significant on the more challenging datasets UNC+ and G-Ref. It leads to 4.12\%, 5.64\% and 2.22\% absolute improvement in overall IoU on the val, testA and testB set of UNC+ and nearly doubles the performance of $prec@0.9$ which requires a precise segmentation mask compared with the groundtruth.  This demonstrates the superior performance of our model for identifying the referred objects with appearance and scene context word only since there is no location word available on the UNC+ dataset. Furthermore, the substantial gain (6.39\% in overall IoU) on the G-Ref dataset proves the advantage of our model in capturing long-range dependencies for cross-modal features and capturing the referred objects based on referring expressions where this dataset contains longer and richer query expressions. The consistent improvement compared with the preliminary work CMSA~\cite{ye2019cross} also proves the word attention-aware pooling in the cross-modal self-attention module can summarizes more effective features from word attentions.

\input tab_miou_gref.tex

\input tab_miou_referit.tex

\subsubsection{Quantitative results of actor and action video segmentation}
We further compare the proposed model for the task of actor and action video segmentation with three baseline methods i.e., referring segmentation model~\cite{hu2016segmentation}, lingual specification model~\cite{li2017tracking} and referring video segmentation model~\cite{gavrilyuk2018actor}.
Table~\ref{tab:miou_a2d} and Table~\ref{tab:miou_jhmdb} present the segmentation results on the A2D Sentences dataset and JHMDB Sentences dataset, respectively. The proposed referring image segmentation model obtains comparable performance on the A2D Sentences dataset and better performance on the JHMDB Sentences dataset compared with the prior best model in \cite{gavrilyuk2018actor}. After adding CFSA to bring temporary information of videos, our model achieves superior segmentation results over almost all evaluation metrics except a slightly inferior result at $prec@0.5$ on the A2D Sentences dataset. 

\input tab_miou_a2d.tex

\input tab_miou_jhmdb.tex

\input fig_seg.tex

\input fig_seg_vid.tex

\input fig_word.tex

\subsubsection{Qualitative results}  
Some qualitative examples generated by our network are shown in Figure~\ref{fig:seg} for referring image segmentation and Figure~\ref{fig:seg_vid} for actor and action video segmentation. To better understand the benefit of multi-level self-attentive features, we visualize the linguistic representation to show attention distributions at different levels by Eq.~\ref{eq:w_att} in Sec.~\ref{sec:msa}.
In this way we can get a vector of length $N$ and repeat this operation for all three levels to finally obtain a matrix of $3\times N$. This matrix is shown in the 2nd column of Fig.~\ref{fig:seg} and the 1st column of Figure~\ref{fig:seg_vid}. We can see that the attention distribution over words corresponding to a particular feature level is different. Features at higher levels (e.g. $l_3$) tend to focus on words that refer to objects (e.g. ``bowl'', ``elephant", ``vase''). Features at lower levels (e.g. $l_1$, $l_2$) tend to focus on words that refer to attributes (e.g. ``white'',``walking") or relationships (e.g. ``left of'', ``middle right",``second'') which helps complement detailed visual cues to localize the object of interest. As shown in the Fig.~\ref{fig:seg}, our model can handle the scenarios that the referred object is at relative position to other objects with the same category, such as the bowl is far left compared with the others in the first example and even a specific order of the vase as ``second from right" in the fourth example. Our model can also identify the particular object which is performing opposite move status to others shown in the second example. In another two interesting examples, the referred object can be successfully identified with discriminated appearance and location words only in the third example and exclude unexpected regions by indicating a negation word ``not".

Some actor and action video segmentation examples are presented in the Figure~\ref{fig:seg_vid}. In the first row, the falling car and the jumping ball have huge variations in terms of shape, location, size, etc. Our model successfully captures these variations. The examples in the middle and bottom rows show the scenarios where there are multiple objects existing in the videos. The proposed approach works well to distinguish different objects of the same (middle examples) or different (bottom examples) categories according to the corresponding descriptions. Surprisingly at the bottom video, when the referred object ``person" disappears in the last frame, our model can still make an ideal prediction that it does not output any segmentation mask.

\input fig_failure.tex

\subsection{Visualization and Failure Cases}\label{sec:vis}

\subsubsection{Visualization}
We visualize spatial feature representations with various query expressions for a given image. This helps to gain further insights on the learned model. 

We adopt the same technique in~\cite{li2018referring} to generate visualization heatmaps over spatial locations. It is created by normalizing the strongest activated channel of the last feature map, which is then upsampled to match with the size of the original input image. Note that here we attempt various referring expressions that are different from those in the test set to show the generalization ability of our model. Fig.~\ref{fig:word} presents these visualized heatmaps. It can be observed that our model is able to correctly respond to different query expressions with various categories, locations and relationships. For instance, it can identify the object of the same category with different noun words like ``chair" and ``couch", and ``crowd" and ``people" in the first two rows. When the query is ``people'' and ``umbrella'', our model can highlight every person and umbrella in the image as shown in the second row. Similarly, in the last row, when the query is ``soccer player'', it captures all players present in the scenario. For more specific phrases such as ``soccer player in red uniform'' and ``soccer player is sliding", the model accurately identifies the correspond players being referred to.

\subsubsection{Failure cases}
We also present some interesting failure examples in Fig.~\ref{fig:failure}. These failure cases are caused by the ambiguity of the language (e.g. two boys on right in the first example), similar object appearance and occlusion (e.g. two giraffes are standing front and back with quite similar appearance). The bottom two rows show three frames of a video and our segmentation masks. Our methods fails to produce the precise segmentation mask because of the disappearing of the fast moving object (e.g., the green car is quickly passing through the camera with varying deformations). Some of these failure cases may potentially be fixed by applying object detectors.

%% file: tab_ab_pool.tex
\begin{table*}[ht!]
\caption{Ablation study of variants for cross-modal self-attention module. The proposed word attention-aware pooling method is compared with different pooling strategies (max pooling and average pooling) to collect the attention distribution of each word. All methods use the same base model (DeepLab-101) without multi-level feature fusion module or DenseCRF for postprocessing.
}
\begin{center}
\scalebox{1.2}{
\begin{tabular}{|c|ccc|ccc|c|c|}
\hline
	 &  &UNC& & &UNC+& & G-Ref  & ReferIt\\

	    & val &  testA &  testB &  val &  testA &  testB & val  & test\\
\hline
    max pooling &    49.90 &  50.91 & 46.34 &  31.85 & 34.66  & 29.07   & 32.96 & 57.92\\ 
    average pooling &  50.12 & 51.22 & 46.85 & 32.14 & 34.98  & 30.31  & 33.41 & 58.53\\
    word attention-aware pooling &   \bf{50.95} &  \bf{51.79} &\bf{47.54} & \bf{34.60}  &  \bf{37.40}   &  \bf{32.02}   & \bf{37.35}  & \bf{59.95}\\

\hline
\end{tabular}
}
\end{center}
\label{tab:ab_pool}
\end{table*}

%% file: tab_ab_att.tex
\begin{table}[ht!]
\caption{Ablation study on the UNC val set. The top five methods evaluate different attention methods and the middle four methods compare results of different feature combination methods for multimodal feature representations. The bottom five results show comparisons of multi-level feature fusion methods. CMSA and GMLF denote the proposed cross-modal self-attention and gated multi-level fusion modules. All methods use the same base model (DeepLab-101) and DenseCRF for postprocessing. 
}
\begin{center}
\scalebox{1.2}{
\begin{tabular}{|c|c|}
\hline
\textbf{Attention Method}  & Overall IoU \\
\hline  
No attention &  45.63    \\
Word attention  & 47.01  \\
Pixel attention   &   47.84 \\ 
Word-pixel pair attention & 47.57  \\
Cross-modal self-attention & {\bf 50.95} \\
\hline

\textbf{Multimodal feature representation} & Overall IoU \\
\hline
    RMI-LSTM~\cite{liu2017recurrent} &  45.18   \\
    RRN-CNN~\cite{li2018referring} &   46.95   \\
    CMSA-S  & 48.53   \\
    CMSA-W  & \bf{50.95}   \\

\hline

\textbf{Multi-level feature fusion} & Overall IoU \\
\hline
    CMSA+PPM  &  53.54   \\
    CMSA+Deconv  &  54.18 \\
    CMSA+ConvLSTM  &  56.56   \\
    CMSA+Gated  &  57.08  \\ 
    CMSA+GMLF (Ours)  & \bf{58.92}  \\

\hline
\end{tabular}
}
\end{center}
\label{tab:ab_att}
\end{table}

%% file: tab_miou_unc.tex
\begin{table*}[ht!]
\caption{Comparison of segmentation performance with the state-of-the-art methods on the UNC dataset in terms of prec@X and IoU.}
\begin{center}
\scalebox{1.15}{
\begin{tabular}{|c|c|ccccc|c|}
\hline
    Method & Set & prec@0.5 & prec@0.6 &  prec@0.7  & prec@0.8 &  prec@0.9 & Overall IoU\\
\hline

RMI~\cite{liu2017recurrent} & val & 42.99  & 33.24  & 22.75  & 12.11 &  2.23    & 45.18 \\
DMN~\cite{margffoy2018dynamic}  & val &  -   & -  & - &  - & -   & 49.78 \\
RRN~\cite{li2018referring}      & val  & 61.66  & 52.50  & 42.40  & 28.13 &   8.51      & 55.33 \\
CMSA~\cite{ye2019cross}         & val  & 66.44 & 59.70  & 50.77  & 35.52 & 10.96 & 58.32 \\
Ours		& val & \textbf{69.24} &  \textbf{62.46} &  \textbf{53.34}      &   \textbf{37.84} &  \textbf{11.99}  & \textbf{58.92}  \\
\hline
RMI~\cite{liu2017recurrent} & testA & 42.99  & 33.59  & 23.69  & 12.94 &  2.44    & 45.69 \\
DMN~\cite{margffoy2018dynamic}  & testA &  65.83   & 57.82  & 46.80 &  27.64 & 5.12   & 54.83 \\
RRN~\cite{li2018referring}      & testA  & 64.13  & 54.66  & 44.37  & 29.15 &   8.08      & 57.26 \\
CMSA~\cite{ye2019cross}      & testA  & 70.02 & 63.71  & 54.64  & 38.04 & 11.65 & 60.61 \\
Ours		& testA & \textbf{73.87} &  \textbf{67.76} &  \textbf{58.53}      &   \textbf{40.52} &  \textbf{12.04}  & \textbf{62.01}  \\

\hline
RMI~\cite{liu2017recurrent} & testB & 44.99  & 34.21  & 22.69  & 11.84 &  2.65    & 45.57 \\
DMN~\cite{margffoy2018dynamic}  & testB &  -   & -  & - &  - &-   & 45.13 \\

RRN~\cite{li2018referring}      & testB  & 59.35  & 50.32  & 39.82  & 27.30 &   10.05      & 53.95 \\
CMSA~\cite{ye2019cross}         & testB  & 61.32 & 53.84  & 44.69  & 32.78 & 12.58 & 55.09 \\
Ours		& testB & \textbf{64.55} &  \textbf{56.62} &  \textbf{47.11}      &   \textbf{34.41} &  \textbf{12.78}  & \textbf{56.03}  \\
\hline
\end{tabular}
}
\end{center}

\label{tab:miou_unc}
\end{table*}

%% file: tab_miou_unc+.tex
\begin{table*}[ht!]
\caption{Comparison of segmentation performance with the state-of-the-art methods on the UNC+ dataset in terms of prec@X and IoU.}
\begin{center}
\scalebox{1.15}{
\begin{tabular}{|c|c|ccccc|c|}
\hline
    Method & Set & prec@0.5 & prec@0.6 &  prec@0.7  & prec@0.8 &  prec@0.9 & Overall IoU\\
\hline

RMI~\cite{liu2017recurrent} & val & 20.52 & 14.02  & 8.46  & 3.77 &  0.62    & 29.86 \\
DMN~\cite{margffoy2018dynamic}  & val &  -   &  - & - & -  &  -  & 38.88 \\
RRN~\cite{li2018referring}      & val  & 37.32 & 28.96  & 20.31  & 11.33 &  2.66       & 39.75 \\
CMSA~\cite{ye2019cross}         & val  & 44.63 & 37.86  & 29.86  & \textbf{20.10} & 5.33 & 43.76 \\
Ours		& val & \textbf{45.48} &  \textbf{38.44} &  \textbf{30.08}      &   19.77 &  \textbf{5.63}  & \textbf{43.87}  \\
\hline
RMI~\cite{liu2017recurrent} & testA & 21.22 & 14.43  & 8.99  & 3.91 &  0.49    & 30.48 \\
DMN~\cite{margffoy2018dynamic}  & testA &  -   &  - & - & -  &  -  & 44.22 \\
RRN~\cite{li2018referring}      & testA  & 40.80 &  31.66 & 22.74  & 12.78 &  2.78       & 42.15 \\
CMSA~\cite{ye2019cross}         & testA  & 51.19 & \textbf{43.42}  & 34.25  & 22.41 & 5.82 & 47.60 \\
Ours		& testA & \textbf{51.41} &  43.19 &  \textbf{34.40}      &   \textbf{23.65} &  \textbf{6.08}  & \textbf{47.79}  \\
\hline
RMI~\cite{liu2017recurrent} & testB & 20.78 & 14.56  & 8.80  & 4.58 &  0.80    & 29.50 \\
DMN~\cite{margffoy2018dynamic}  & testB &  -   &  - & - & -  &  -  & 32.29 \\
RRN~\cite{li2018referring}      & testB  & 32.42 & 24.69  & 17.10  & 9.92 & 2.78   & 36.11 \\
CMSA~\cite{ye2019cross}         & testB  & 35.02 &  28.68 & 21.74  & 13.89 & 4.54 & 37.89 \\
Ours		& testB & \textbf{37.57} &  \textbf{30.89} &  \textbf{23.95}      &   \textbf{15.32} &  \textbf{5.26}  & \textbf{38.33}  \\
\hline
\end{tabular}
}
\end{center}

\label{tab:miou_unc+}
\end{table*}

%% file: tab_miou_gref.tex
\begin{table*}[ht!]
\caption{Comparison of segmentation performance with the state-of-the-art methods on the G-Ref dataset in terms of prec@X and IoU.}
\begin{center}
\scalebox{1.2}{
\begin{tabular}{|c|c|ccccc|c|}
\hline
    Method & Set & prec@0.5 & prec@0.6 &  prec@0.7  & prec@0.8 &  prec@0.9 & Overall IoU\\
\hline

RMI~\cite{liu2017recurrent} & val & 27.77 & 21.06  & 13.92  & 6.83 & 1.43     & 34.52 \\
DMN~\cite{margffoy2018dynamic}  & val &  -   & -  & - & -  & -   & 36.76 \\
KWA~\cite{shi2018key}  & val & 27.85 & 21.01  & 13.42   & 6.60 & 1.97    & 36.92  \\
RRN~\cite{li2018referring}      & val  & 36.00 & 29.77  & 22.78  & 14.06 &  3.74       & 36.45 \\
CMSA~\cite{ye2019cross}         & val  & 40.38 & 32.56  & 24.77  & 16.10 & 4.59 & 39.98 \\
Ours		& val & \textbf{44.71} &  \textbf{37.26} &  \textbf{29.10}      &   \textbf{19.02} &  \textbf{5.59}  & \textbf{42.84}  \\
\hline
\end{tabular}
}
\end{center}
\label{tab:miou_gref}
\end{table*}

%% file: tab_miou_referit.tex
\begin{table*}[ht!]
\caption{Comparison of segmentation performance with the state-of-the-art methods on ReferIt dataset in terms of prec@X and IoU.}
\begin{center}
\scalebox{1.2}{
\begin{tabular}{|c|c|ccccc|c|}
\hline
    Method & Set & prec@0.5 & prec@0.6 &  prec@0.7  & prec@0.8 &  prec@0.9 & Overall IoU\\
\hline

RMI~\cite{liu2017recurrent} & test & 46.08 & 38.90  & 30.77  & 20.62 &  8.54    & 58.73 \\
DMN~\cite{margffoy2018dynamic}  & test &  -   & -  & - & -  &  -  & 52.81 \\
KWA~\cite{shi2018key}  & test & 45.87 & 39.80  &  32.82  & 23.81 & 11.79     & 59.09 \\
RRN~\cite{li2018referring}      & test  & 56.71 & 49.22  & 40.36  & 28.39 &  12.68     & 63.63 \\
CMSA~\cite{ye2019cross}        & test  & 57.96 &  50.34 &  41.04 & 28.82 & 13.05 & 63.80 \\
Ours		& test & \textbf{58.25} &  \textbf{51.04} &  \textbf{42.12}      &   \textbf{29.96} &  \textbf{13.25}  & \textbf{63.99}  \\
\hline
\end{tabular}
}
\end{center}

\label{tab:miou_referit}
\end{table*}

%% file: tab_miou_a2d.tex
\begin{table*}[ht!]
\caption{Comparison of segmentation performance with the state-of-the-art methods on A2D Sentences dataset in terms of prec@X, meanIoU and Overall IoU.}
\begin{center}
\scalebox{1.2}{
\begin{tabular}{|c|ccccc|cc|}
\hline
    Method &  prec@0.5 & prec@0.6 &  prec@0.7  & prec@0.8 &  prec@0.9 & Overall IoU  & Mean IoU\\
\hline
Hu \emph{et al.}~\cite{hu2016segmentation} & 34.8 & 23.6 & 13.3 & 3.3  &0.1  &  47.4    & 35.0 \\
Li \emph{et al.}~\cite{li2017tracking}  & 38.7 & 29.0 & 17.5 & 6.6  & 0.1 & 51.5     & 35.4 \\
Gavrilyuk \emph{et al.}~\cite{gavrilyuk2018actor}  & \textbf{50.0} & 37.6 & 23.1 & 9.4  & 0.4 &  55.1 & 42.6 \\
Ours  & 46.7 & 38.5 & 27.9 & 13.6  & 1.7 &  59.2 & 40.5 \\
Ours+CFSA	& 48.7 & \textbf{43.1} &  \textbf{35.8} &  \textbf{23.1}      &   \textbf{5.2} &  \textbf{61.8}  &  \textbf{43.2} \\
\hline
\end{tabular}
}
\end{center}
\label{tab:miou_a2d}
\end{table*}

%% file: tab_miou_jhmdb.tex
\begin{table*}[ht!]
\caption{Comparison of segmentation performance with the state-of-the-art methods on JHMDB Sentences dataset in terms of prec@X, Overall IoU and Mean IoU.}
\begin{center}
\scalebox{1.2}{
\begin{tabular}{|c|ccccc|cc|}
\hline
    Method &  prec@0.5 & prec@0.6 &  prec@0.7  & prec@0.8 &  prec@0.9 & Overall IoU  & Mean IoU\\
\hline
Hu \emph{et al.}~\cite{hu2016segmentation} & 63.3  & 35.0 & 8.5 & 0.2  & 0.0 & 54.6 & 52.8 \\
Li \emph{et al.}~\cite{li2017tracking}  & 57.8 & 33.5 & 10.3 & 0.6  & 0.0 &  52.9  & 49.1 \\
Gavrilyuk \emph{et al.}~\cite{gavrilyuk2018actor}  & 69.9 & 46.0 & 17.3 & 1.4  & 0.0 &  54.1    & 54.2 \\
Ours  & 76.0 & 60.4 & 36.4 & 6.4  & 0.0 &  60.6    & 57.0 \\
Ours+CFSA	&  \textbf{76.4} & \textbf{62.5} &  \textbf{38.9} &  \textbf{9.0}      &   \textbf{0.1} &  \textbf{62.8}  &  \textbf{58.1} \\
\hline
\end{tabular}
}
\end{center}
\label{tab:miou_jhmdb}
\end{table*}

%% file: fig_seg.tex
\begin{figure*}[!htb]
  \centering
{\setlength{\tabcolsep}{0.5pt}
\scalebox{0.8}{
  \begin{tabular}{cccc}

      \multicolumn{4}{c}{Query:  \textit{``the bowl of food to the left of all the other bowls''}}\\
     \includegraphics[width=1.2in]{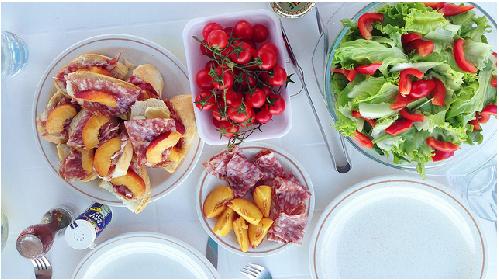}&
     \includegraphics[height=0.7in]{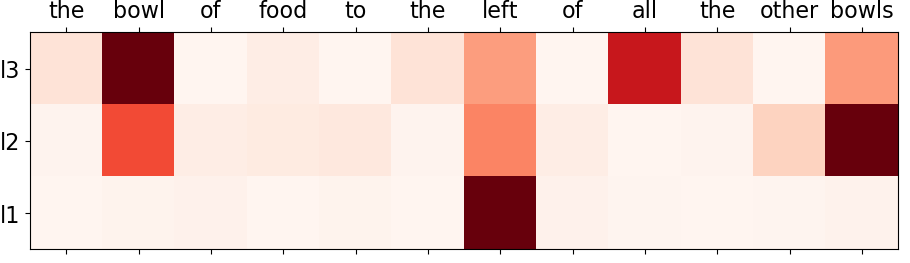}&
      \includegraphics[width=1.2in]{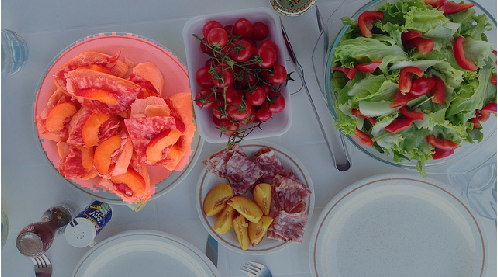}&
      \includegraphics[width=1.2in]{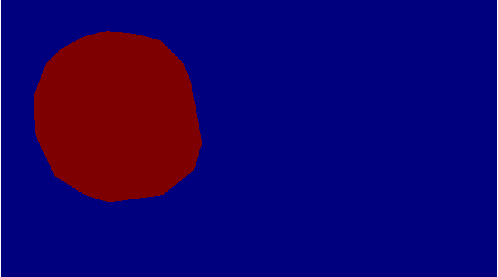}\\

       \multicolumn{4}{c}{Query:  \textit{``an elephant walking opposite direction of the others''}} \\
    \includegraphics[width=1.2in]{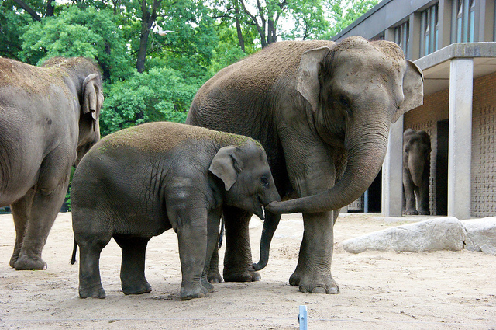}&
    \includegraphics[height=0.8in]{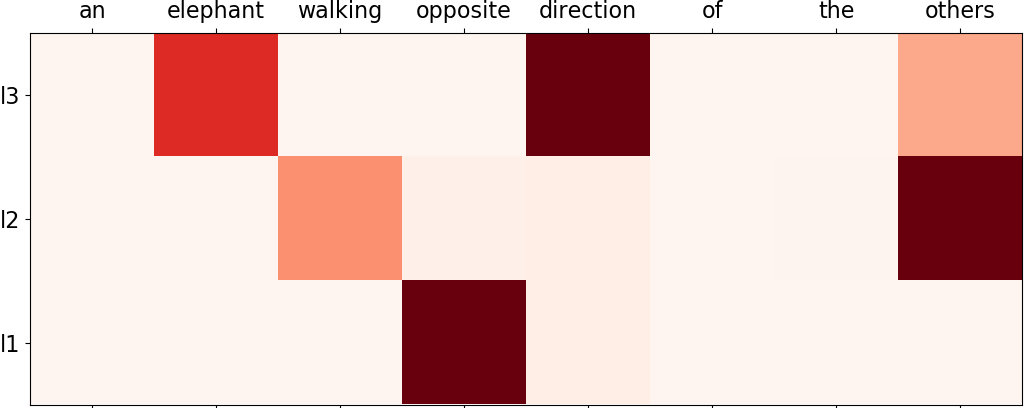}&
     \includegraphics[width=1.2in]{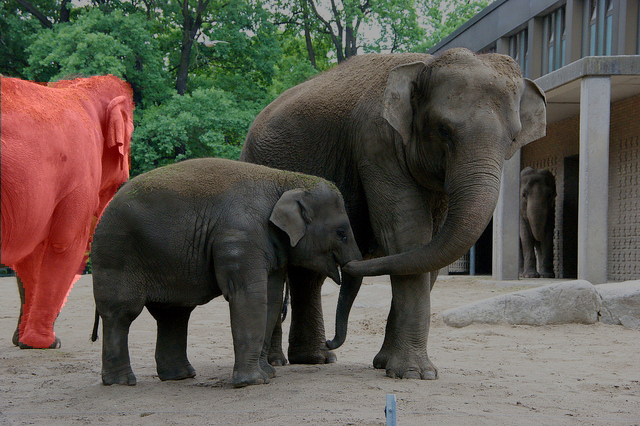}&
      \includegraphics[width=1.2in]{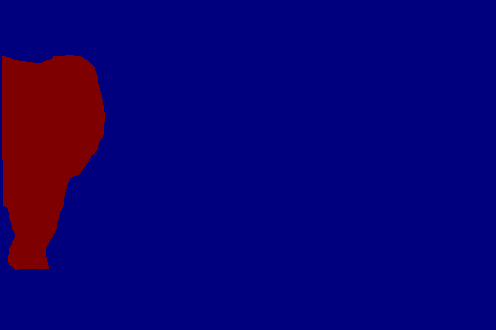}\\
       
            \multicolumn{4}{c}{Query:  \textit{``white face middle right''}} \\
    \includegraphics[width=1.2in]{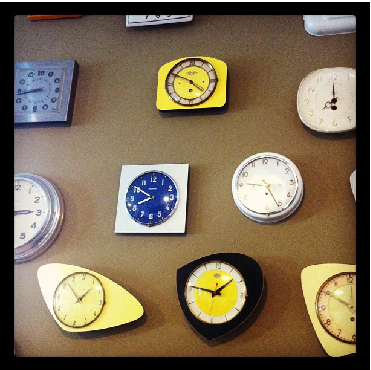}&
    \includegraphics[height=1.2in]{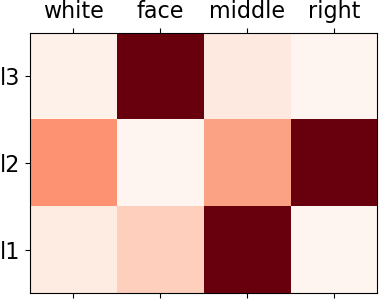}&
     \includegraphics[width=1.2in]{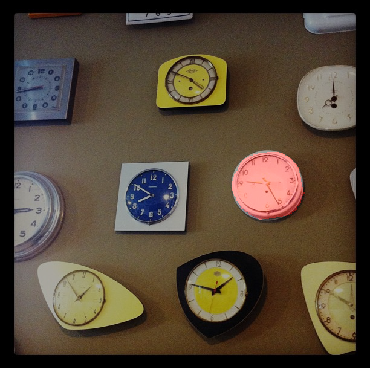}&
      \includegraphics[width=1.2in]{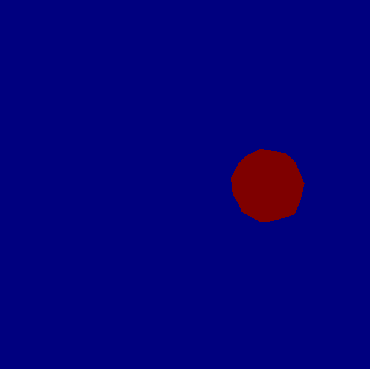}\\

    \multicolumn{4}{c}{Query:  \textit{``second vase from right''}} \\
    \includegraphics[width=1.2in]{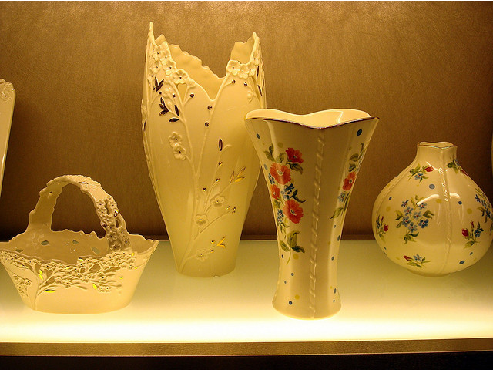}&
    \includegraphics[height=0.9in]{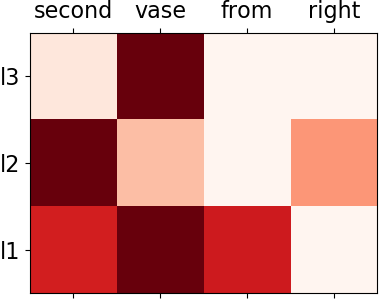}&
     \includegraphics[width=1.2in]{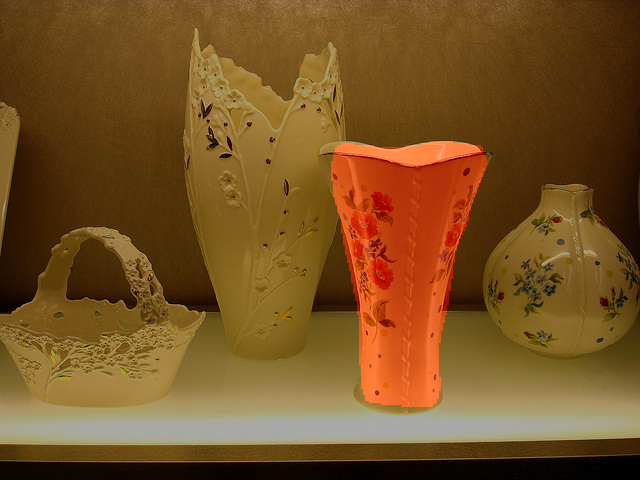}&
      \includegraphics[width=1.2in]{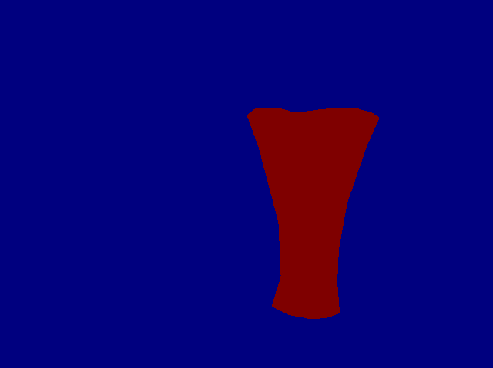}\\     
        
         \multicolumn{4}{c}{Query:  \textit{``sky, not the clouds''}} \\
    \includegraphics[width=1.2in]{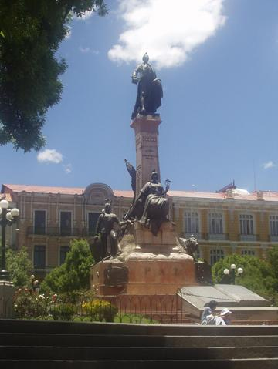}&
    \includegraphics[height= 1.6in]{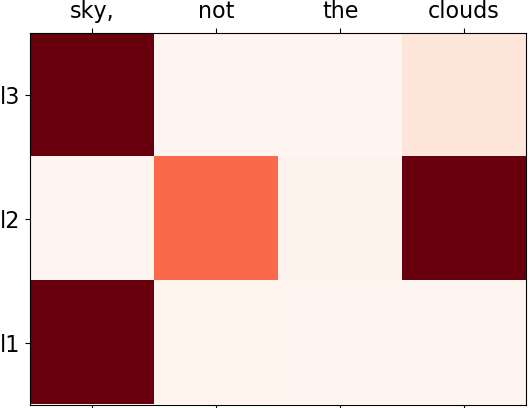}&
     \includegraphics[width=1.2in]{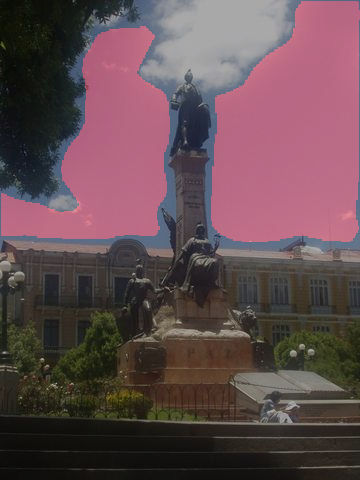}&
      \includegraphics[width=1.2in]{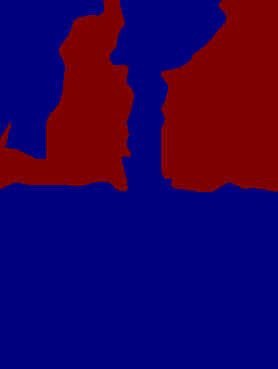}\\ 
      
        (a) & (b) & (c) &(d)
        \\
  \end{tabular}
}
}
\caption{Qualitative examples of referring image segmentation: (a) original image; (b) visualization of the linguistic representation (attentions to word at each of the three feature levels); (c) segmentation mask and; (d) ground truth.}
\label{fig:seg}
\end{figure*}

%% file: fig_seg_vid.tex
\begin{figure*}[!htb]
  \centering
{\setlength{\tabcolsep}{0.5pt}
\scalebox{0.78}{
  \begin{tabular}{cccccc}

      \multicolumn{3}{c}{Query:  \textit{``big black car is falling down the hill''}}&\multicolumn{3}{c}{Query:  \textit{``an orange ball is jumping up and down''}}\\
     \includegraphics[width=1.3in]{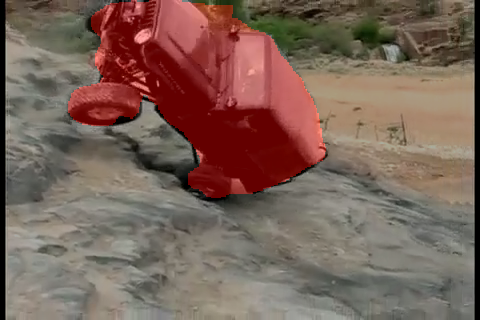}&
     \includegraphics[width=1.3in]{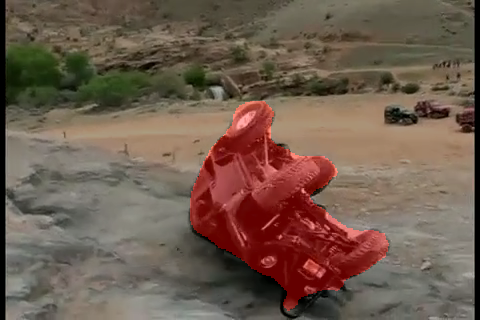}&
      \includegraphics[width=1.3in]{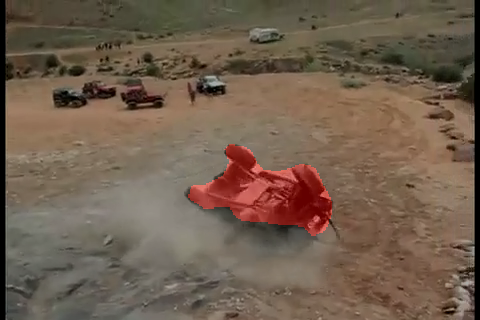}&
            \includegraphics[width=1.3in, height=0.87in]{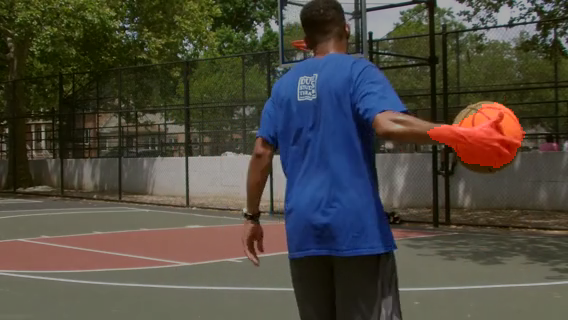}&
        \includegraphics[width=1.3in, height=0.87in]{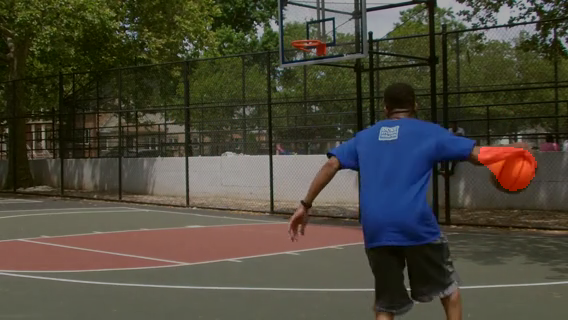}&
      \includegraphics[width=1.3in, height=0.87in]{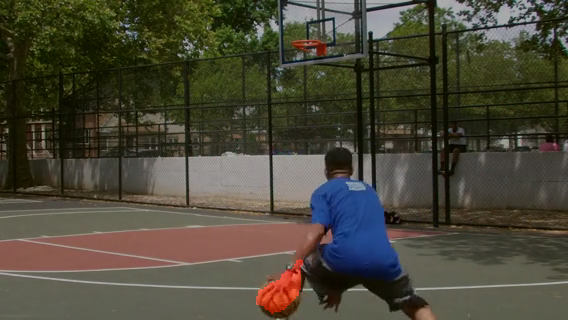}\\

      \multicolumn{3}{c}{Query:  \textit{``a black cat is sitting on the back''}}&\multicolumn{3}{c}{Query:  \textit{``a gray white cat on the left is approaching''}}\\
 \includegraphics[width=1.3in]{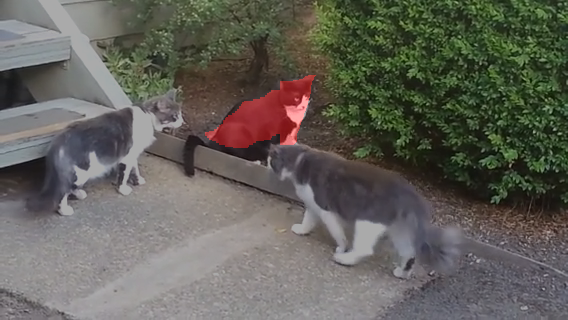}&
     \includegraphics[width=1.3in]{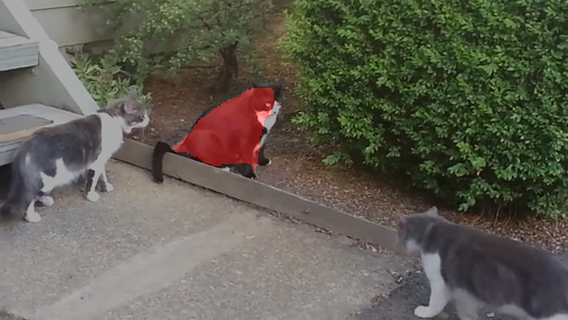}&
      \includegraphics[width=1.3in]{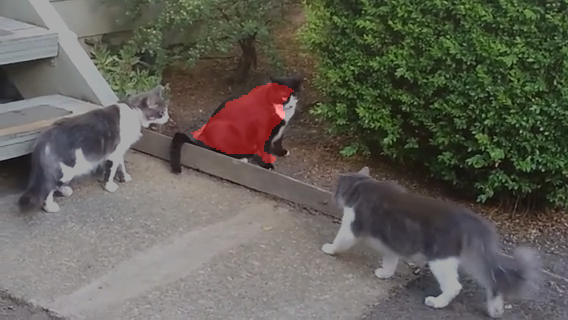}&
            \includegraphics[width=1.3in]{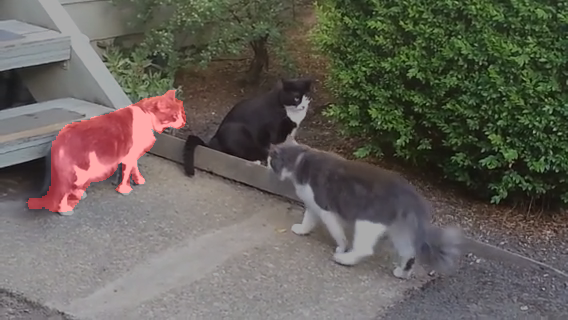}&
        \includegraphics[width=1.3in]{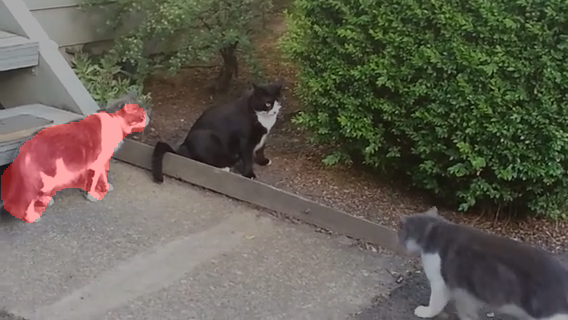}&
      \includegraphics[width=1.3in]{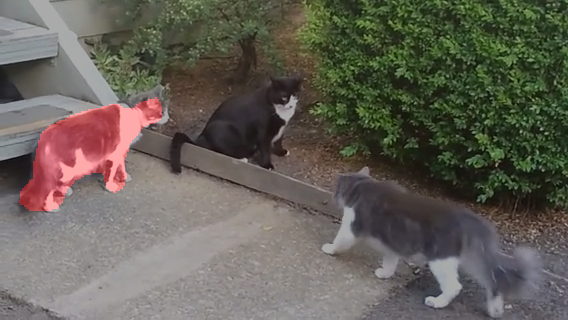}\\

           \multicolumn{3}{c}{Query:  \textit{``white fuzzy dog is running in the corridor''}}&\multicolumn{3}{c}{Query:  \textit{``person walking the dog''}}\\
     \includegraphics[width=1.3in]{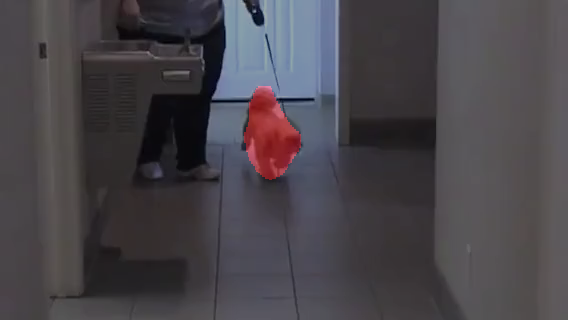}&
     \includegraphics[width=1.3in]{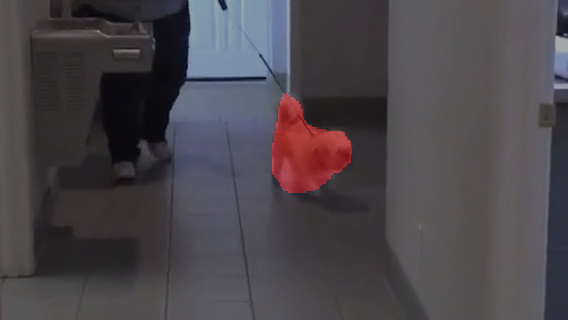}&
      \includegraphics[width=1.3in]{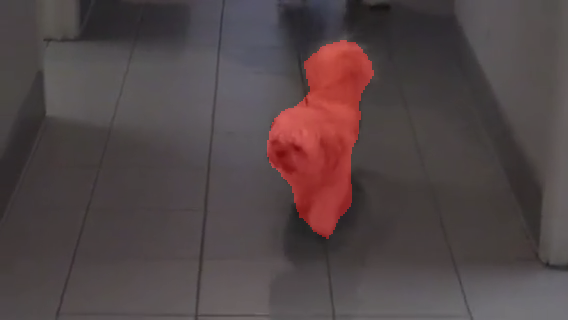}&
            \includegraphics[width=1.3in]{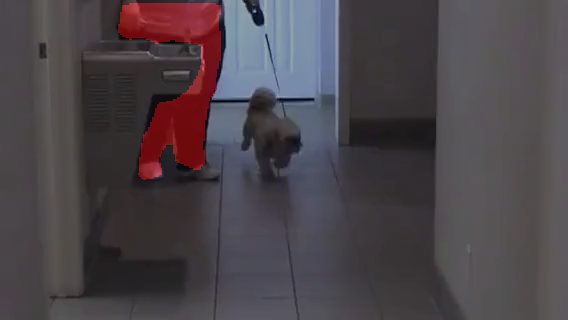}&
        \includegraphics[width=1.3in]{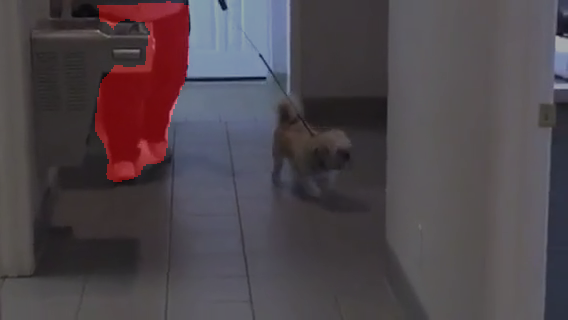}&
      \includegraphics[width=1.3in]{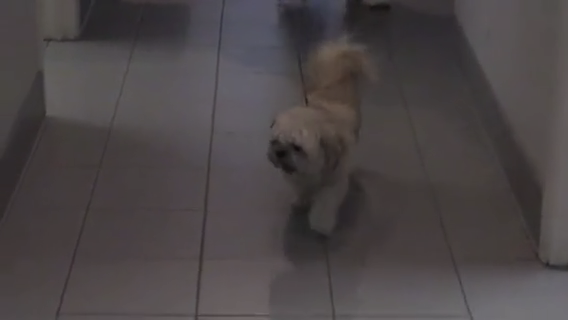}\\

  \end{tabular}
}
}

\caption{Qualitative examples of actor and action video segmentation. From left to right: the first three and last three examples represent segmentation masks for different frames in a video according to the query expressions.}
\label{fig:seg_vid}
\end{figure*}

%% file: fig_word.tex
\begin{figure*}[!htb]
  \centering
{
\scalebox{0.85}{
  \begin{tabular}{ccccc}

      \includegraphics[width=1.21in]{figure/Fig_vis/6915_im.png}&
    \includegraphics[width=1.21in]{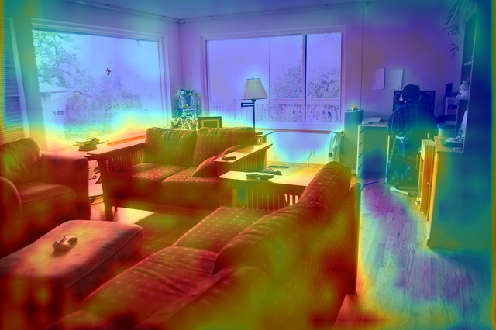}&
    \includegraphics[width=1.21in]{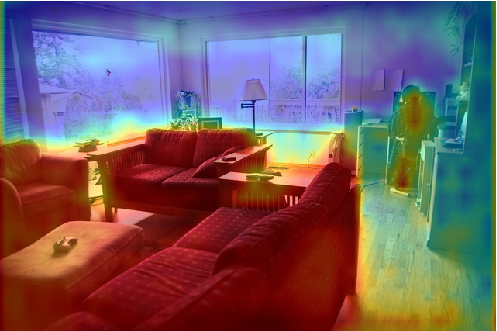}&
    \includegraphics[width=1.21in]{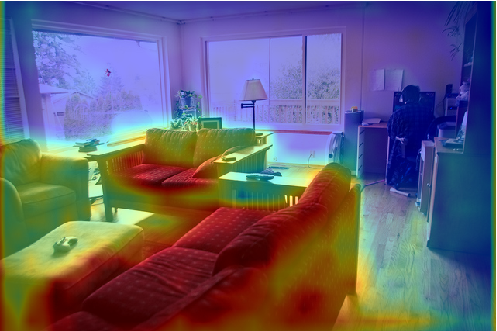}& 
     \includegraphics[width=1.21in]{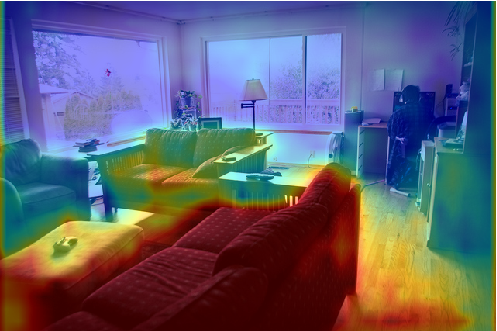}\\
       & \textit{``chair"} & \textit{``couch"} & \textit{``red couch"} &  \textit{``the red couch} \\ &&&&       \textit{closest to you"} \\ 
     
           \includegraphics[width=1.21in]{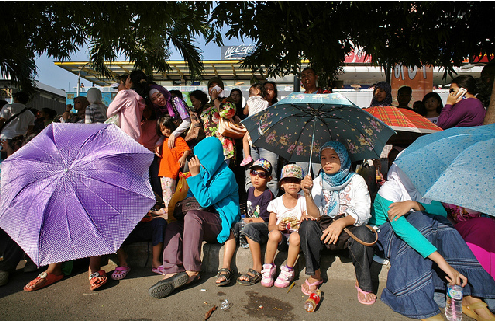}&
    \includegraphics[width=1.21in]{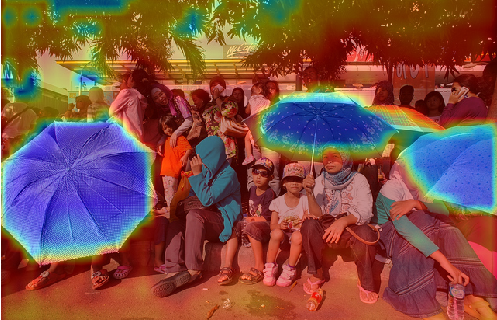}&
    \includegraphics[width=1.21in]{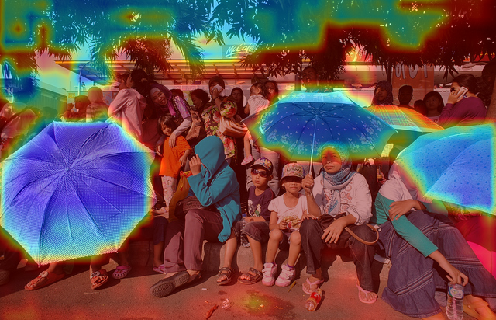}&
    \includegraphics[width=1.21in]{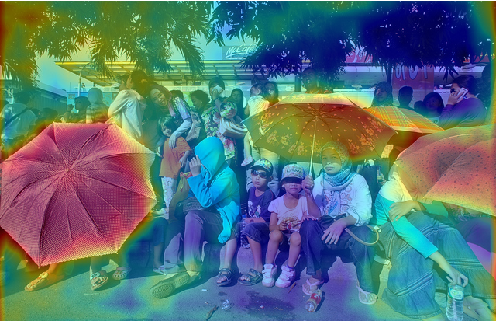}& 
     \includegraphics[width=1.21in]{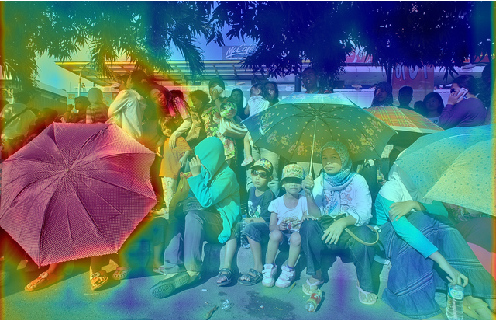}\\
        & \textit{``crowd"} & \textit{``people"} & \textit{``umbrella"} &  \textit{``a purple umbrella"} \\     
      
    \includegraphics[width=1.21in]{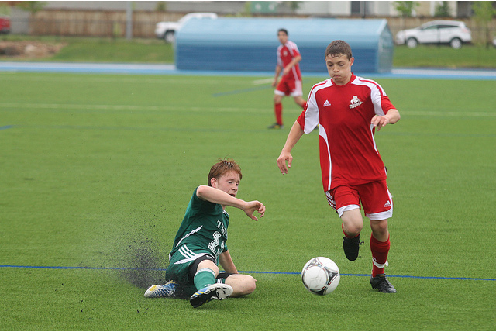}&
    \includegraphics[width=1.21in]{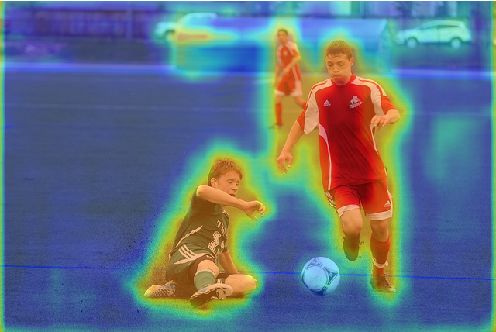}&
    \includegraphics[width=1.21in]{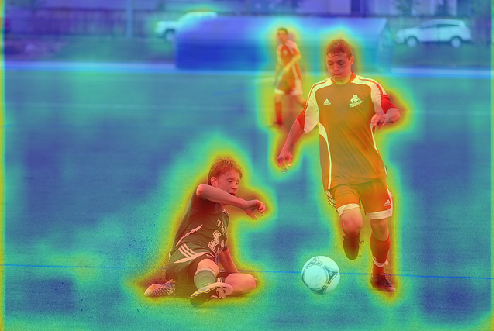}&
    \includegraphics[width=1.21in]{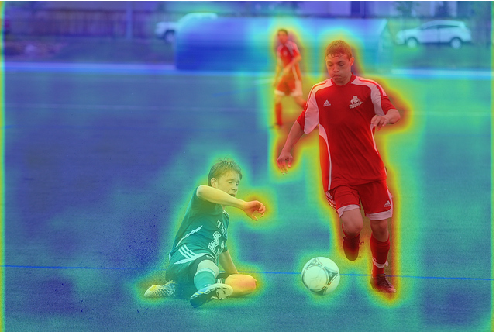}& 
     \includegraphics[width=1.21in]{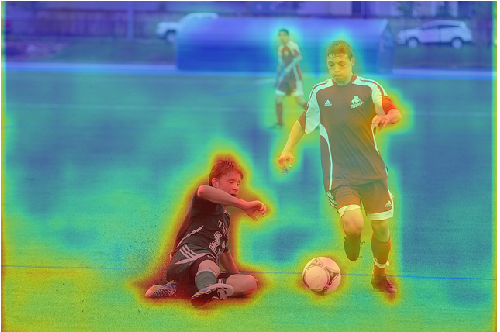}\\
    
      & \textit{``soccer player"} & \textit{``soccer player} & \textit{``soccer player} &  \textit{``soccer player} \\   & &\textit{on the grass"}&  \textit{in red uniform"}  & \textit{ is sliding"}  \\
      
 \end{tabular}
}
}
\caption{(Best viewed in color) Visualization of spatial feature representation. These spatial heatmaps show the responses of the network to different query expressions.}
\label{fig:word}
\end{figure*}

%% file: fig_failure.tex
\begin{figure}[htb]
  \centering
{\setlength{\tabcolsep}{0.5pt}
\scalebox{0.76}{
  \begin{tabular}{ccc}

         \multicolumn{3}{c}{Query: \textit{``boy on right''}}    \\
    \includegraphics[width=1.2in]{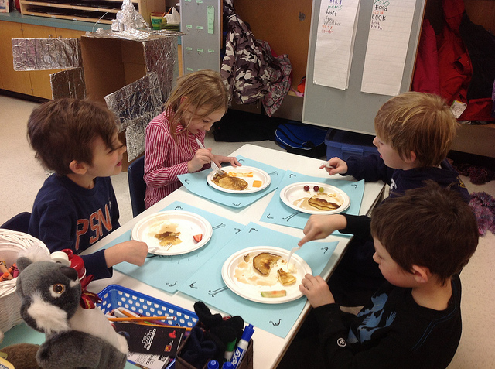}&
    \includegraphics[width=1.2in]{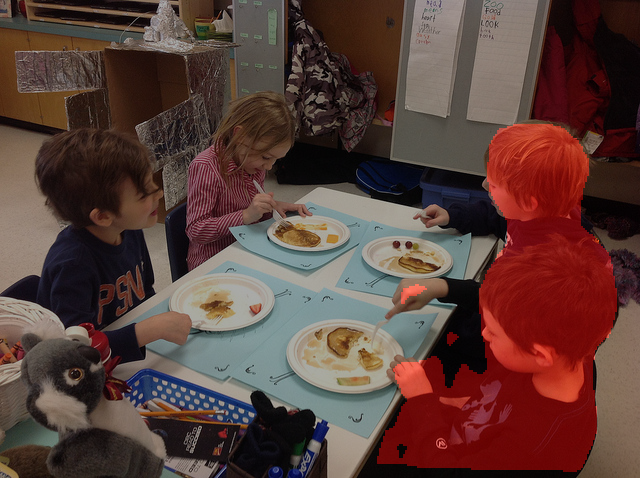}&
    \includegraphics[width=1.2in]{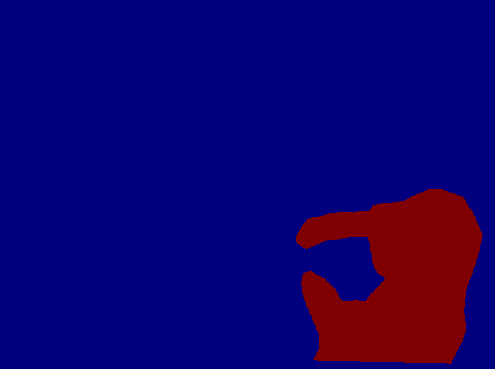}\\

       \multicolumn{3}{c}{Query:  \textit{``giraffe standing next to a giant giraffe''}}    \\  
    \includegraphics[width=1.2in]{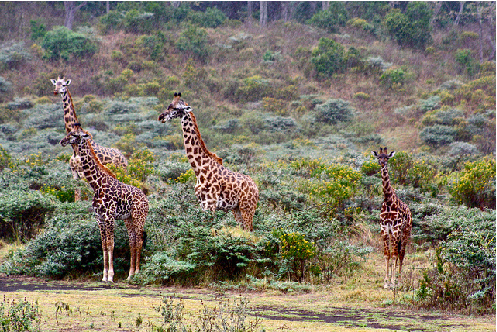}&
    \includegraphics[width=1.2in]{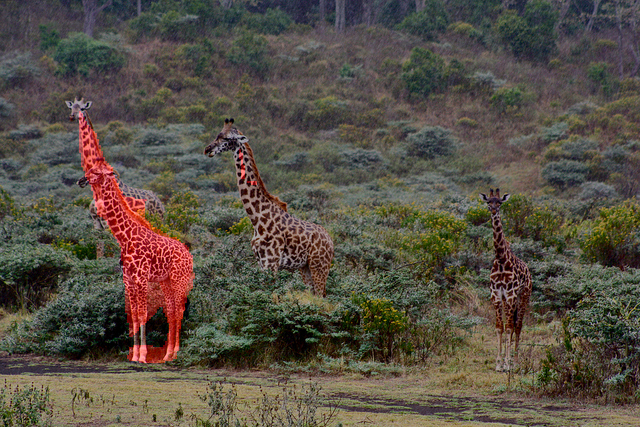}&
    \includegraphics[width=1.2in]{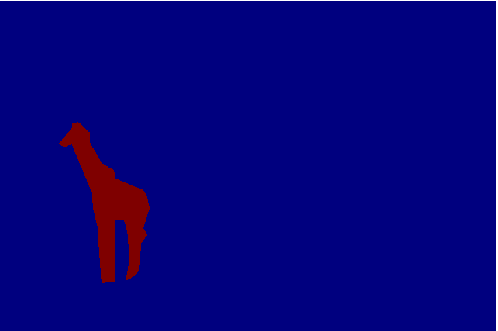}\\
 
        \multicolumn{3}{c}{Query:  \textit{``green car running on the street''}}    \\  
    \includegraphics[width=1.18in]{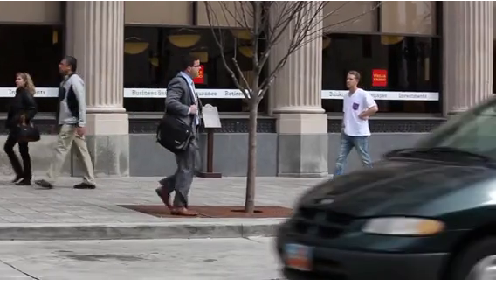}&
    \includegraphics[width=1.18in]{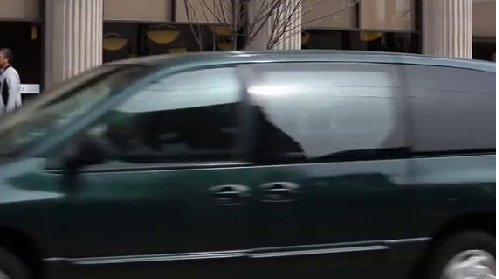}&
    \includegraphics[width=1.18in]{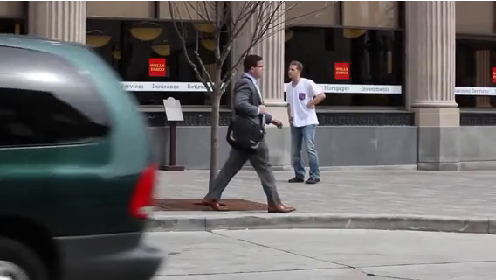}\\
   
    \includegraphics[width=1.18in]{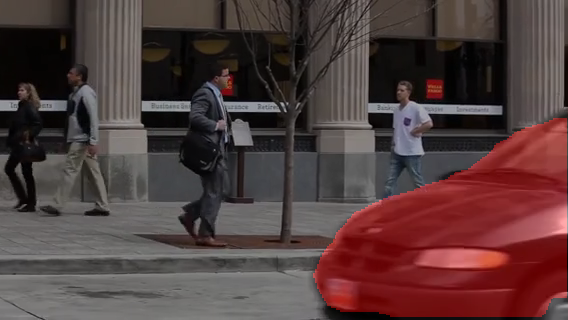}&
    \includegraphics[width=1.18in]{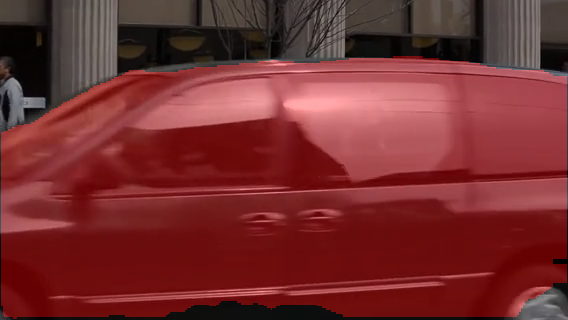}&
    \includegraphics[width=1.18in]{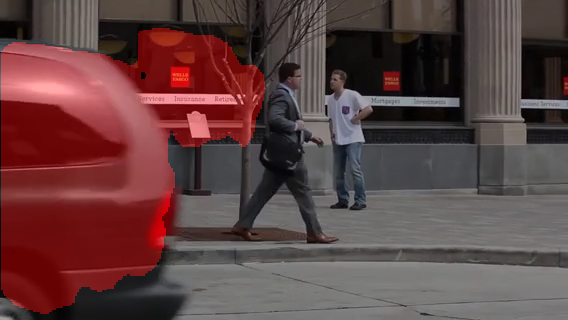}\\

  \end{tabular}
}
}
\caption{Some failure examples of our model. In the first two rows, we show examples of failure cases in images. Here we show the original image (1st column), the output of our method (2nd column) and the ground-truth (3rd column). The failures of images are due to factors such as language ambiguity (1st row), and similar object appearance and occlusion (2nd row). We also show a failure case of video in the last two rows. Row 3 shows frames in a video and row 4 shows the corresponding outputs of our method. Our model fails to precisely segment the object in the last frame because of the disappearing of the fast moving object.}
\label{fig:failure}
\end{figure}

%% file: conclude.tex
\section{Conclusion}\label{sec:conclude}
We have proposed cross-modal self-attention and gated multi-level fusion modules to address two crucial challenges in the referring segmentation task, and cross-frame self-attention module to extend our model to video domain. Our cross-modal self-attention module captures long-range dependencies between visual and linguistic modalities, which results in a better feature representation to focus on important information for referred entities. The proposed gated multi-level fusion module adaptively integrates features from different levels via learnable gates for each individual level. In addition, the cross-frame self-attention module effectively extract  temporal correlation feature from adjacent frames for the current frame. The proposed network achieves state-of-the-art results on all four referring image benchmark datasets and two actor and action video segmentation datasets.